\pdfoutput=1

\documentclass[11pt]{article}

\usepackage{acl}

\usepackage{times}
\usepackage{latexsym}

\usepackage{comment}
\usepackage{rotating}
\usepackage[utf8]{inputenc}
\usepackage{times}
\usepackage{latexsym}
\usepackage{rotating,tabularx,siunitx}

\usepackage{float}
\usepackage{graphicx}
\usepackage{subfigure}
\usepackage{caption}
\usepackage{amsmath,amssymb}
\usepackage{mathtools}
\usepackage{booktabs}
\usepackage{multirow}
\usepackage{multicol}
\usepackage{siunitx}
\usepackage{adjustbox}
\usepackage{pifont}

\usepackage{enumitem}

\usepackage{color}

\usepackage[T1]{fontenc}

\usepackage[utf8]{inputenc}

\usepackage{microtype}
\newcommand*{\affaddr}[1]{#1} 
\newcommand*{\affmark}[1][*]
{\textsuperscript{#1}}

\title{DiscoPrompt: Path Prediction Prompt Tuning for Implicit Discourse Relation Recognition}
\author{
Chunkit Chan\thanks{\quad Equal contribution.}\ \ \affmark[1],
Xin Liu$^{*}$\affmark[1],
Jiayang Cheng\affmark[1],
Zihan Li\affmark[1],
Yangqiu Song\affmark[1],\\
\textbf{Ginny Y. Wong\affmark[2],
Simon See\affmark[2]}\\
\affaddr{\affmark[1]Department of Computer Science and Engineering, HKUST, Hong Kong SAR, China}\\
\affaddr{\affmark[2]NVIDIA AI Technology Center (NVAITC), NVIDIA, Santa Clara, USA} \\
\texttt{\{ckchancc, xliucr, jchengaj, zliho, yqsong\}@cse.ust.hk} \\
\texttt{\{gwong, ssee\}@nvidia.com}\\
}

\begin{document}
\maketitle

\begin{abstract}

Implicit Discourse Relation Recognition (IDRR) is a sophisticated and challenging task to recognize the discourse relations between the arguments with the absence of discourse connectives. The sense labels for each discourse relation follow a hierarchical classification scheme in the annotation process~\cite{DBLP:conf/lrec/PrasadDLMRJW08}, forming a hierarchy structure. Most existing works do not well incorporate the hierarchy structure but focus on the syntax features and the prior knowledge of connectives in the manner of pure text classification. We argue that it is more effective to predict the paths inside the hierarchical tree (e.g., ``\textit{Comparison} -> \textit{Contrast} -> \textit{however}'') rather than flat labels (e.g., \textit{Contrast}) or connectives (e.g., \textit{however}). We propose a prompt-based path prediction method to utilize the interactive information and intrinsic senses among the hierarchy in IDRR. This is the first work that injects such structure information into pre-trained language models via prompt tuning, and the performance of our solution shows significant and consistent improvement against competitive baselines.

\end{abstract}
\section{Introduction}
Discourse parsing is the task of automatically parsing discourse structure in a text, including the identification of discourse structure and the annotation of discourse relations~\cite{DBLP:journals/fcsc/LiLQL22}. Discourse Relation Recognition (DRR) is a crucial task in discourse parsing, recognizing relations between two arguments (i.e., sentences or clauses). It is vital for textual coherence and is considered as the essential step for many downstream tasks involving more context, such as question answering~\cite{DBLP:conf/naacl/RutherfordX15}, text generation~\cite{DBLP:conf/naacl/BosselutCHGHC18}, and argument mining~\cite{DBLP:conf/acl/LiuOSJ20}. Explicit discourse relation recognition (EDRR) has already been demonstrated that utilizing explicit connectives information can effectively determine the discourse relation types~\cite{DBLP:conf/sigdial/VariaHC19}. On the other hand, implicit discourse relation recognition (IDRR) is still challenging with the absence of connectives~\cite{DBLP:conf/sigdial/VariaHC19}. 

\begin{figure}[t]
\small
\centering
\includegraphics[width=\linewidth]{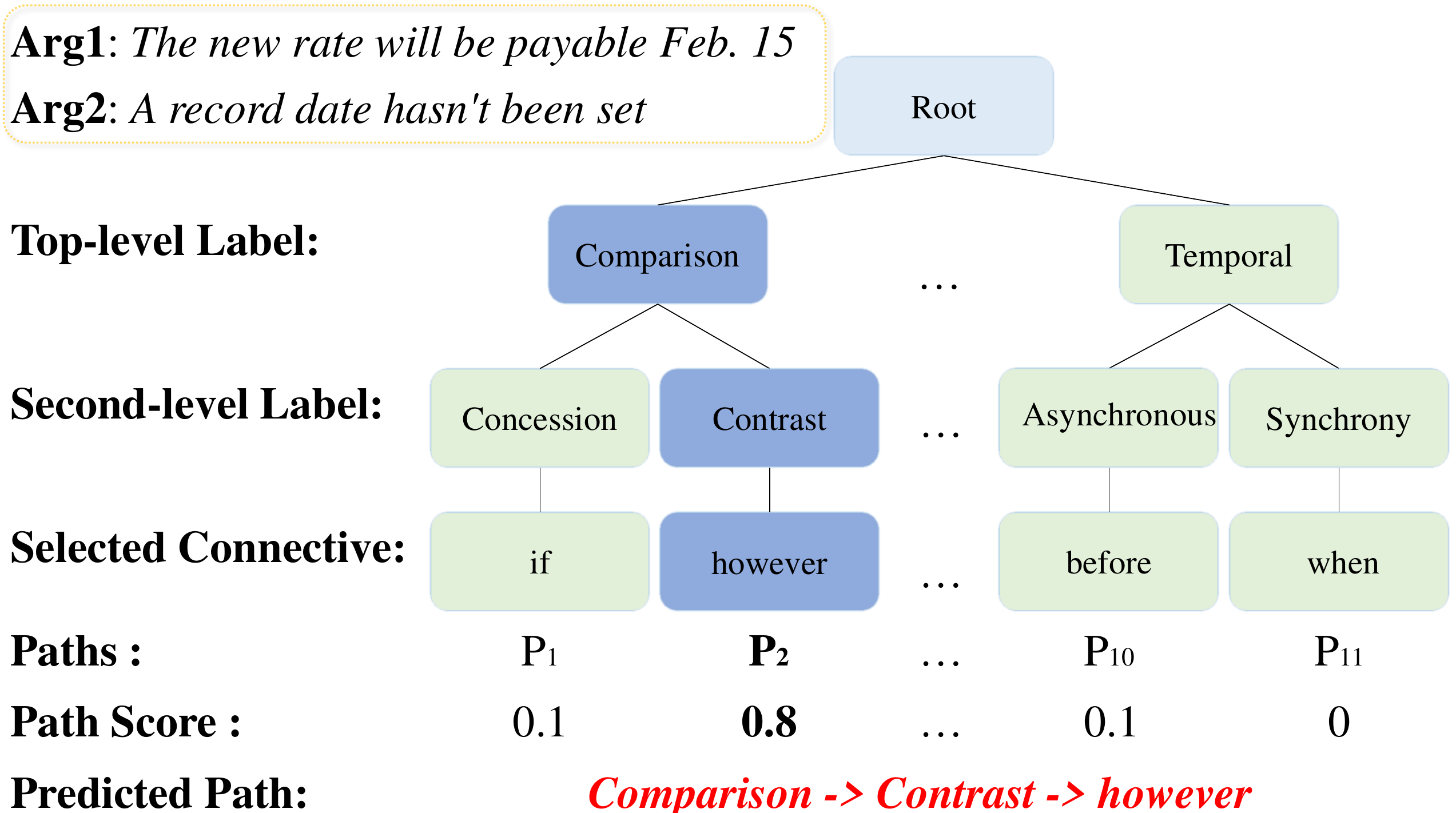}
\vspace{-0.2in}
\caption{
An example of the implicit discourse relation hierarchy and path prediction.
}
\label{fig:DiscoPrompt_PathScore}
\vspace{-0.2in}
\end{figure}

Traditional works on IDRR focus on syntax features, including word pairs~\cite{DBLP:conf/emnlp/LinKN09, DBLP:conf/sigdial/VariaHC19} and other surface features~\cite{DBLP:journals/tacl/JiE15, Bai2018Deep}.
With deep neural networks and large language models (LLMs), different approaches pay much attention to text representations via attention~\cite{Liu2016Recognizing}, pre-training~\cite{Shi2019Next}, multi-task learning~\cite{DBLP:conf/acl/HeWGH20,conf/emnlp/Contrastive_Learning}, and prior knowledge~\cite{DBLP:conf/ijcai/LiuOSJ20,DBLP:journals/corr/abs-2210-07032}.
But one important piece of information, i.e., the inherent discourse label hierarchy, is not fully investigated.

The sense labels for each discourse relation follow a hierarchical classification scheme in the annotation process of PDTB 2.0 framework~\cite{DBLP:conf/lrec/PrasadDLMRJW08}, forming a hierarchy structure.
Figure~\ref{fig:DiscoPrompt_PathScore} shows an example from PDTB 2.0 dataset~\cite{DBLP:conf/lrec/PrasadDLMRJW08}.
It consists of two arguments (i.e., Arg1 and Arg2) and is annotated with relation senses, where the semantics of the top-level \textit{Comparison} is further refined by the second-level \textit{Contrast}.
Besides, we list representative connectives (e.g., \textit{however}) to help better understand the definitions and semantics of labels.
LDSGM~\cite{DBLP:conf/aaai/WuCGLZS22} uses graph convolutional networks to encode the label dependencies into text representations, illustrating the importance of label structures on text representation learning and label prediction.
However, such usage is not compatible with pre-training because it may significantly affect the representations from language models. 
Prompt tuning has shown its power in text classification without altering the representations from pre-trained language models, especially for low-resource scenarios~\cite{DBLP:conf/eacl/SchickS21, DBLP:conf/acl/GaoFC20}.

In this paper, we propose a prompt-based path prediction method, \textbf{Disco}urse relation path prediction \textbf{Prompt} tuning model (\textbf{DiscoPrompt}~\footnote{The source code is available at~\url{https://github.com/HKUST-KnowComp/DiscoPrompt}}), to utilize the hierarchy and intrinsic senses of labels in IDRR.
Specifically, we transform the hierarchy in Figure~\ref{fig:DiscoPrompt_PathScore} to ``Comparison -> Concession -> if; $\cdots$; Temporal -> Synchrony -> when'' as the hierarchical prompt and add it as the prefix of arguments to be classified.
The dependencies of top and second-level relation senses are explicitly provided as the context.
On the other hand, connectives are provided as the natural language explanations of labels to help the language models better adapt to the prior knowledge. 
We ask the LLMs to predict the label's hierarchical path instead of the leaf label for IDDR, and we show such a way of providing the label hierarchy ahead of arguments significantly improves the IDRR performance.
Our contributions are summarized as follows:
{\begin{itemize}[leftmargin=*]
    \item This is the first work that injects labels' hierarchical structure information and connectives into pre-trained language models via prompt tuning.
    \item We model the IDRR problem as the path prediction problem that predicts the joint probability of top-level relations, second-level types, and connectives at the same time.
    \item We conduct extensive experiments and thorough ablation studies to discuss the necessity and effectiveness of the label hierarchy and connectives. The results support our claims and the success of our proposed DiscoPrompt model.
\end{itemize}}

\section{Related Work}
\paragraph{Prompt Tuning}
With LLMs, such as T5~\cite{DBLP:journals/jmlr/RaffelSRLNMZLL20} and GPT-3~\cite{DBLP:conf/nips/BrownMRSKDNSSAA20}, prompt-based methods have attracted much attention in the field of natural language understanding~\citep{DBLP:conf/eacl/SchickS21,DBLP:conf/emnlp/LesterAC21,DBLP:conf/acl/0010LLWWBCH22}. Compared with fine-tuning, prompt tuning may have a better generalization on various tasks due to the aligned nature of language descriptions and answer semantics, e.g., classification problems~\cite{DBLP:conf/acl/GaoFC20,DBLP:conf/naacl/WangXM22}.
At the same time, there are some efforts to leverage prompts with structural inputs for knowledge customization~\citep{DBLP:conf/naacl/ZhongGDQL0W0D22}.
Injecting hierarchy information into prompts is also promising. For example, using top-level predictions to refine prompts of bottom levels can surpass soft prompts and hard prompts~\cite{DBLP:journals/corr/abs-2204-13413}. Nevertheless, how to employ LLMs to better involve hierarchy knowledge is still under investigation.

\paragraph{Implicit Discourse Relation Recognition}
It has been discovered that connectives can provide necessary clues in predicting discourse relations to achieve around 95\% accuracy \cite{Dai2019A, DBLP:conf/sigdial/VariaHC19}. However, the absence of connectives makes the prediction more challenging.
Many efforts have been paid to explore the syntax through linguistic features~\cite{DBLP:conf/naacl/RutherfordX15,DBLP:journals/tacl/JiE15,DBLP:conf/conll/WangL16,DBLP:conf/naacl/DaiH18,DBLP:conf/sigdial/VariaHC19}, attention~\cite{Liu2016Recognizing,Bai2018Deep}, pre-training~\cite{Shi2019Next}, knowledge transfer~\cite{Lan2017Multi,Dai2019A,DBLP:conf/acl/HeWGH20}, etc.
With the power of language models, connective prediction also illustrates its effectiveness in implicit relation prediction~\cite{DBLP:conf/acl/NguyenLTN19,DBLP:conf/iwcs/ShiD19,DBLP:conf/lrec/KishimotoMK20,DBLP:journals/corr/abs-2106-03192}.
In addition, PCP~\cite{DBLP:journals/corr/abs-2210-07032} shows the feasibility of combining label prediction and connective prediction under the manner of prompts.
The latest methods reveal the significance of the label hierarchy of discourse relations. LDSGM~\cite{DBLP:conf/aaai/WuCGLZS22} utilizes the graph convolutional networks to incorporate label dependencies into text representations, while ContrastiveIDRR~\cite{conf/emnlp/Contrastive_Learning} leverages the sense hierarchy to obtain contrastive learning representation. 
However, these methods are incompatible with pre-training as they modify the representations from pre-trained language models. 
Therefore, this work investigates injecting the label dependencies information and connectives into pre-trained language models via prompt tuning with aligning the representations.


\section{Method}
\begin{figure}[t]
    \vspace{-0.25in}
    \centering
    \includegraphics[width=\linewidth]{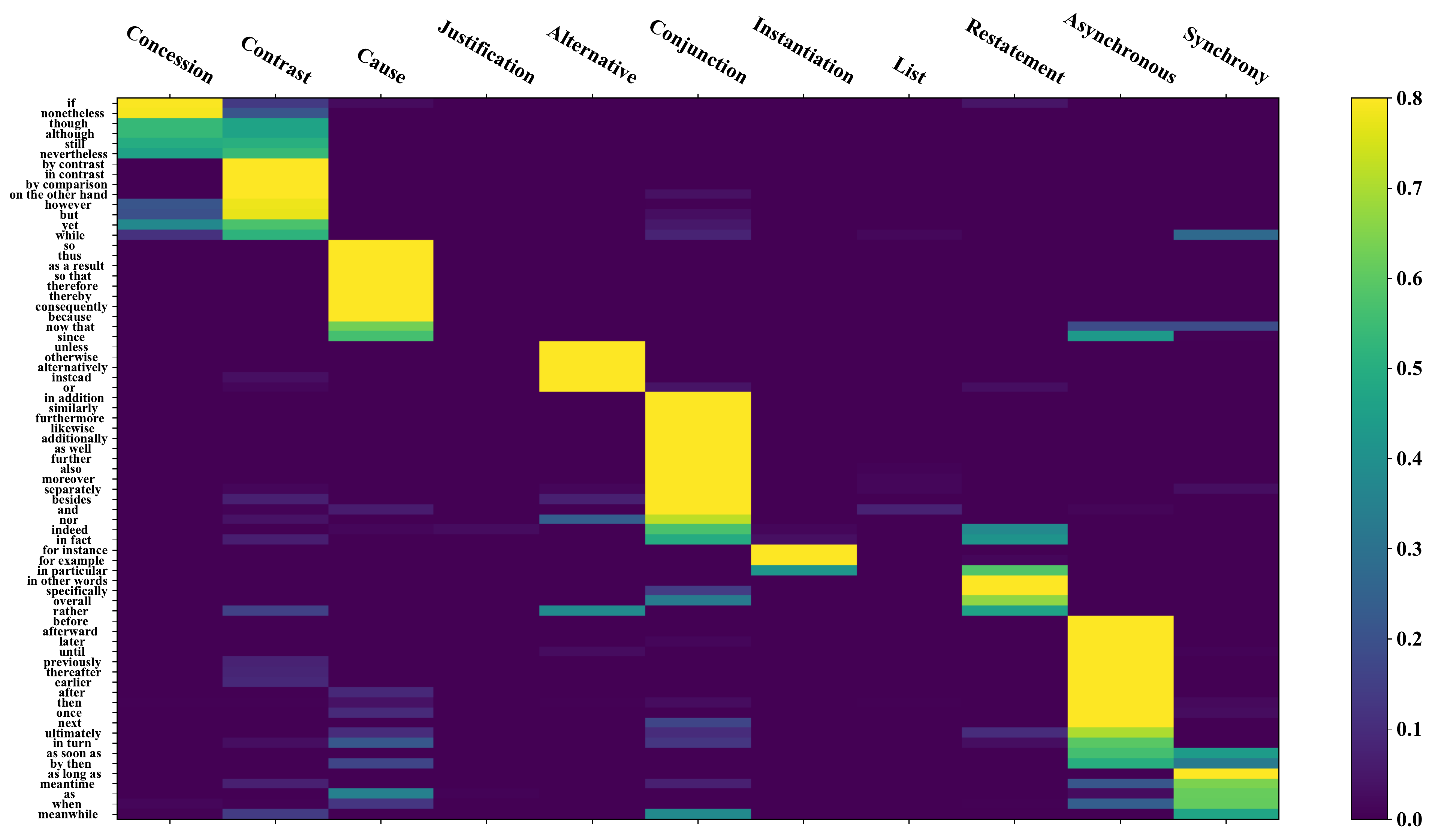}
    \vspace{-0.3in}
    \caption{Prior probabilities of PDTB 2.0 frequent connectives.
    }
    \label{fig:pdtb_heatmap}
    \vspace{-0.2in}
\end{figure}

\subsection{Problem Definition} 
The sense labels in various levels of the Implicit Discourse Relation Recognition (IDRR) task naturally constitute a hierarchy, denoted as $\mathcal{H}$.
$\mathcal{H}$ is a hierarchical tree structure whose depth is $d$, with the root node in depth $0$ and class sense of different levels distributed to the corresponding layer (i.e., from depth $1$ to $d$) in this tree.
Let class label set ${\mathcal{C}}$ to be $\bigcup_{k=1}^{d} \mathcal{C}^k$ where $\mathcal{C}^k = \{c_{1}^k, \cdots, c_{n_k}^k\}$ is the label set of depth $k$, and $n_k$ is the number of classes at depth $k$.
For example, the hierarchy $\mathcal{H}$ of PDTB 2.0 forms a tree with depth size 2, and the $\mathcal{C}^2$ corresponds to the label set of the second level, containing 11 class subtypes like \textit{Concession}, \textit{Synchrony}, etc.
We can enrich the label hierarchy by adding a connective layer like in Figure~\ref{fig:DiscoPrompt_PathScore}.
We adopt the Naive Bayes to compute the prior distribution $\text{Pr}( c^2 | z)$ from the explicit relation data, where $c^2 \in \mathcal{C}^2$ is a subtype, and $z$ is the connectives.
Figure~\ref{fig:pdtb_heatmap} shows the heat map of highly frequent connectives.
We can find that the connectives are the vital clue for discourse relations.
Therefore, we select the most discriminative ones as $\mathcal{C}^3$.
We do not observe significant improvement when adding more than one connective for each $c^2$.
Therefore, we summarize $\mathcal{C}$ for PDTB 2.0 in Table~\ref{tab:pdtb2_label}.
Prior distributions and label words of CoNLL16 are shown in Appendix~\ref{sec:appendix_discoPrompt_implementation_details}.

In this task, given a data set $\mathcal{D} = \{(x_i,y_i)\}$ consisting of data instance ${x_i} = {(a_i^1,a_i^2)}$ and label ${y_i}$, where the ${a_i^1, a_i^2}$ represent the argument 1 and argument 2 of respective instance $i$ and the label ${y_i}$ is class label set. In our method, the class label set including $d$ labels for $d$ layers forms a path $\mathcal{P}$ in hierarchical tree $\mathcal{H}$, instead of a single class label for a specific level. After predicting a path, the classes of various levels are the nodes lying in the predicted path. 
Therefore, this task is to find out the optimal path:
\scalebox{0.91}{\parbox{1.10\linewidth}{
\begin{align}
    {P}_{i}^{*}=\arg \max _{\mathcal{P}^{j}} \text{Pr}\left(\mathcal{P}^{j}  \mid x_i\right),
\end{align}
}}
where ${P}_{i}^{*}$ is the optimal path and $j$ indicates the $j$-th path among all paths.

\begin{table}[t]
\centering
\vspace{-0.25in}
\small
  \scalebox{0.8}{\begin{tabular}{c| c | c}
  \toprule
  \textbf{Top-level} &  \textbf{Second-level} &  \textbf{Connectives}\\
  \midrule
  \multirow{2}{*}{Comparison} & Concession & if \\
      & Contrast & however\\
  \midrule
  \multirow{2}{*}{Contingency} & Cause & so \\
        & Justification & indeed\\
  \midrule
  \multirow{5}{*}{Expansion} & Alternative & instead \\
                             & Conjunction & also \\ 
                             & Instantiation & for example \\ 
                             & List & and \\ 
                             & Restatement & specifically\\                              
  \midrule
  \multirow{2}{*}{Temporal} & Asynchronous & before \\
                            & Synchrony & when \\
  \bottomrule
  \end{tabular}}
  \caption{
  \label{tab:pdtb2_label}
The label word set on PDTB 2.0 dataset, includes four top-level relations, 11 second-level subtypes, and 11 connectives.
}
\vspace{-0.7cm}
\end{table}

\subsection{T5 Backbone Model}
T5~\cite{DBLP:journals/jmlr/RaffelSRLNMZLL20} is an encoder-decoder model pre-trained on a multi-task mixture of unsupervised and supervised tasks. The unsupervised denoising training task required the model only to predict the masked consecutive spans of tokens. For example, the input “Thank you for inviting me to your party last week.” will be corrupted as “Thank you <X> me to your party <Y> week.” and the target is “ <X> for inviting <Y> last </s>” </s> is the eos\_token. In the supervised pre-trained task, the model was asked to perform the sequence-to-sequence input-output mapping by specifying the task prefix (such as “translate German to English:" or “summarize:”). However, the specific textual prefix token is difficult to discover and requires a substantial amount of human effort. Hence, prefix tuning~\cite{DBLP:conf/acl/LiL20} and prompt tuning~\cite{DBLP:conf/emnlp/LesterAC21} methods proposed to overcome this problem by relaxing the constraint of discrete textual tokens to continuous tunable ones. 

\subsection{Path Prediction Prompt Tuning Method}
\begin{figure*}[t]
\vspace{-0.5cm}
\centering
\includegraphics[width= 0.8\textwidth]{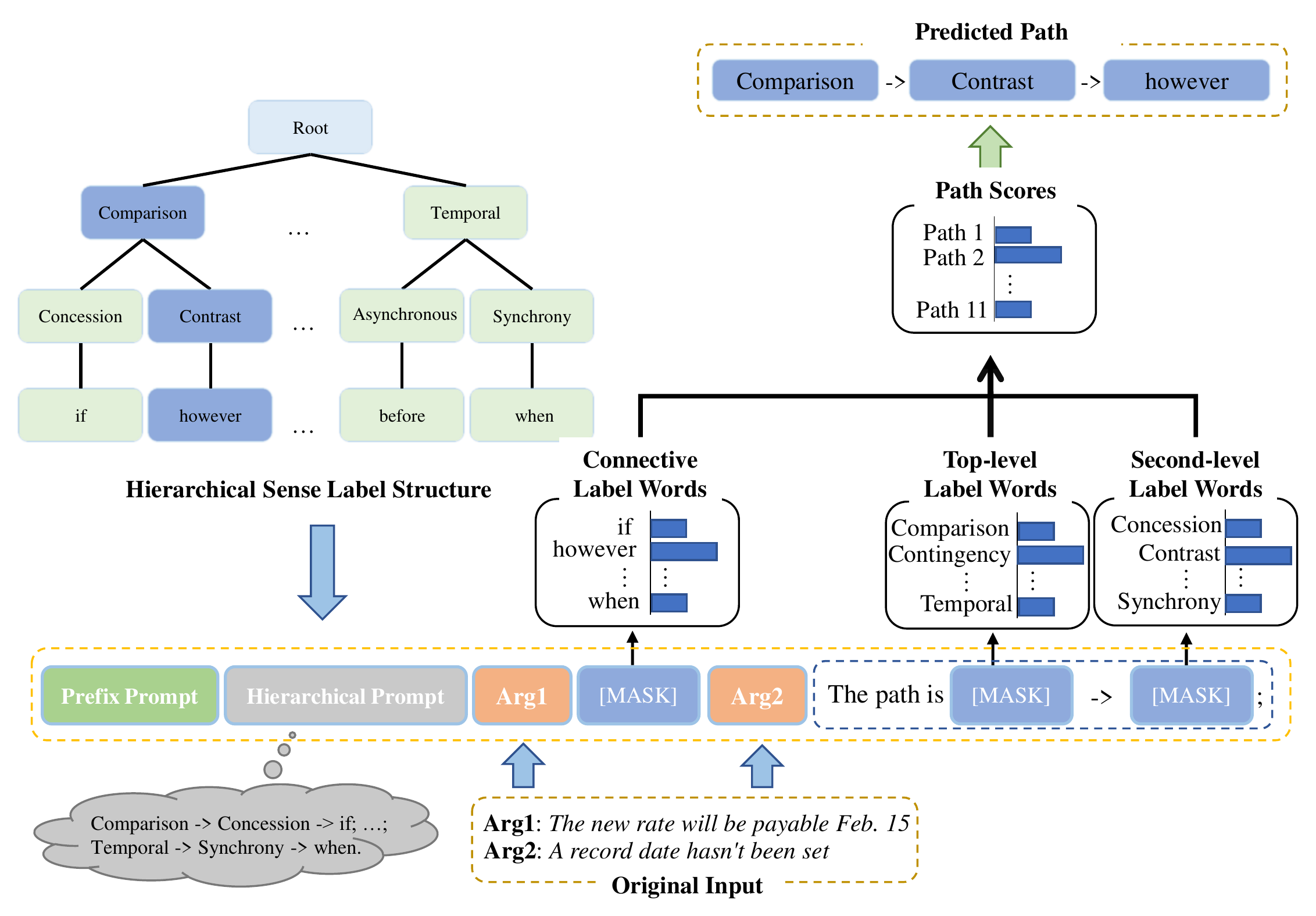}
\vspace{-0.5cm}
\caption{
DiscoPrompt model architecture.
}
\label{fig:DiscoursePrompt_Architecture}
\vspace{-0.25in}
\end{figure*}

To predict the path ${P}_{i}^{*}$ for each instance $x_i$, we leverage a template $\mathcal{T}(\cdot)$ to convert the data instances to a human-tailored template and a verbalizer $\mathcal{V}(\cdot)$ to map a set of words to class labels. 
The template translates the $x_i$ to the prompt input $\tilde{x}_i = \mathcal{T}(x_i)$, and the verbalizer translates.
Figure~\ref{fig:DiscoursePrompt_Architecture} illustrates the archtecture of \textbf{DiscoPrompt}.

\paragraph{Structure-Aware Prompt}

The crafted template includes necessary discrete tokens, masked tokens, soft continuous tokens, and context with the hierarchy information.
The first part of our prompt template is the discrete tokens ``The path is '' for eliciting the predicted path $\mathcal{P}_i$.
Then three [MASK] tokens are included: an [MASK] is inserted between two arguments for predicting the probability of decided connectives, and two [MASK]s form an edge ``[MASK] -> [MASK]'' for receiving the top and second-level class probabilities.
We also added 20 learnable continuous tokens at the beginning of the template to effectively searching an optimal template.
To better utilize the hierarchy information and senses of labels,  we explicitly translate them into a tailored hierarchical tree prompt and insert it into the input.
This hierarchical tree prompt is the discrete tokens appended ahead of the arguments as the context in natural language. Figure~\ref{fig:DiscoPrompt_Template_Searching} in Appendix~\ref{sec: Appendix_ablation_study_DiscoPrompt} shows the details of the template.

\paragraph{Path Verbalizer}
A traditional verbalizer usually maps a label $y$ to a single answer token $z$ or a series of spans $z^1, z^2, \cdots $ greedily ~\cite{DBLP:conf/eacl/SchickS21,DBLP:journals/corr/abs-2107-13586}.
We extend it by mapping a path $\mathcal{P}$ to three tokens, i.e. $\{\mathcal{P}^{j}\} \rightarrow \mathcal{Z} \times \mathcal{Z} \times \mathcal{Z}$, where $\mathcal{Z}$ is the vocabulary.
We denote the three [MASK] tokens as $z^1$, $z^2$, and $z^3$.
Then using the prompt template with three [MASK]s and the verbalizer $\mathcal{V}(\cdot)$, the probability distribution over $\{\mathcal{P}^{j}\}$ can be formalized as the joint probabilities of $z^1$, $z^2$, and $z^3$,  i.e. $\text{Pr}(\mathcal{P}^j \mid \tilde{x}_i) = \text{Pr}(\mathcal{V}(\mathcal{P}^j) \mid \tilde{x}_i) = \text{Pr}(z_i^1 = p_3^j, z_i^2 = p_1^j, z_i^3 = p_2^j \mid \tilde{x}_i)$, where a path $\mathcal{P}^j$ consists of $p_1^j$ (the top-level), $p_2^j$ (the second-level), and $p_3^j$ (the connective).
Since T5 can synchronously predict masked tokens, the joint probability can be written as
\scalebox{0.91}{\parbox{1.10\linewidth}{
\begin{align}
    \text{Pr}(\mathcal{P}^j \mid \tilde{x}_i) = \prod_{k=1}^{3}\text{Pr}(z_i^k = v^k(\mathcal{P}^j) \mid \tilde{x}_i), \label{eq:joint_prob}
\end{align}
}}
where $v^k(\cdot): \{\mathcal{P}^{j}\} \rightarrow \mathcal{Z}$ is the submap of $\mathcal{V}(\cdot)$ for the $k$-th [MASK].
The final learning objective of DiscoPrompt is to maximize
\scalebox{0.91}{\parbox{1.10\linewidth}{
\begin{align}
    \mathcal{J} = \frac{1}{|\mathcal{D}|} \sum_{(x_i, y_i) \in \mathcal{D}} \log \sum_{k=1}^{3} \text{Pr}(z_i^k = v^k(\mathcal{P}^j) \mid \tilde{x}_i).
\end{align}
}
}

Once we get the prediction of $\mathcal{P}_i^{*}$ by choosing the maximum joint probability (i.e., path score) as Eq.~(\ref{eq:joint_prob}), we can get the prediction of each level as Eq.~(\ref{eq:node_prob}).
\scalebox{0.91}{\parbox{1.10\linewidth}{
\begin{align}
    c_i^{k^{*}} = \arg \max_{c^k} \text{Pr}(c^k  \mid \mathcal{P}^j, x_i) \cdot \text{Pr}(\mathcal{P}^j  \mid x_i), \label{eq:node_prob}
\end{align}
}}
where $\text{Pr}(c^k  \mid \mathcal{P}^j, x_i)$ can be calculated by the prior probability (or simply set as 1.0).

\section{Experimental Setting}

\begin{table*}[!ht]
\centering
\vspace{-0.5cm}
\small
  \scalebox{0.9}{\begin{tabular}{ l| cc | cc | cc | cc }
    \toprule
    \multirow{2}{*}{ \textbf{Models} } & \multicolumn{2}{c|}{\textbf{Ji (Top)}} & \multicolumn{2}{c|}{\textbf{Ji (Second)}} & \multicolumn{2}{c|}{\textbf{Lin (Top)}} & \multicolumn{2}{c}{\textbf{Lin (Second)}}  \\
    { } & {\textbf{F1}} & {\textbf{Accuracy}} & {\textbf{F1}}& {\textbf{Accuracy}} & {\textbf{F1}} & {\textbf{Accuracy}} & {\textbf{F1}} & {\textbf{Accuracy}} \\
    \midrule
    MTL-MLoss~\cite{DBLP:conf/acl/NguyenLTN19}  & 53.00 & - & - & 49.95& - & - & - & 46.48\\
    ELMo-C\&E~\cite{Dai2019A}  & 52.89 & 59.66 & 33.41 & 48.23  & - & - & - & -\\
    RWP-CNN~\cite{DBLP:conf/sigdial/VariaHC19} & 50.20 & 59.13 & - & - & - & - & - & -\\ 
    TransS~\cite{DBLP:conf/acl/HeWGH20}  & - & - & - & - & 51.24 & 59.94 & - & -\\ 
    BMGF-RoBERTa~\cite{DBLP:conf/ijcai/LiuOSJ20}  & 63.39 & 69.06 & 35.25 & 58.13 & \textit{58.54} & \textit{68.66} & \textit{39.15} & \textit{53.96}\\    
    CG-T5~\cite{DBLP:conf/emnlp/JiangFCLZ21}  & 57.18 & 65.54 & 37.76 & 53.13 & - & - & - & - \\
    LDSGM~\cite{DBLP:conf/aaai/WuCGLZS22}  & 63.73 &  71.18 & 40.49 & 60.33 & - & - & - & - \\
    GOLF~\cite{DBLP:journals/corr/abs-2211-13873} & 65.76 & 72.52 & 41.74 & 61.16 & - & - & - & -  \\
    ContrastiveIDRR~\cite{conf/emnlp/Contrastive_Learning} & \underline{67.85} &  71.70 & \underline{45.54} & 59.19 & - & - & - & - \\  
    \midrule
    XLNet (base, cased)~\cite{DBLP:conf/acl/KimFGL20}   & \textit{59.33} & \textit{66.35} & \textit{36.36} & \textit{54.73} & \textit{56.16} & \textit{68.05} & \textit{36.23} & \textit{55.82}\\
    XLNet (large, cased)~\cite{DBLP:conf/acl/KimFGL20}  & \textit{63.58} & \textit{69.52} & \textit{38.24} & \textit{61.29} & \textit{58.97} & \textit{\underline{72.17}} & \textit{40.71} & \textit{58.77}\\
    OTMT (XLNet-base)~\cite{DBLP:conf/www/JiangQL22}  &  60.78 & 68.89 & - & 56.65 & - & - & - & 56.37\\
    OTMT (XLNet-large)~\cite{DBLP:conf/www/JiangQL22}  & 64.46 & 72.34 & - & 61.06 & - & - & - & \underline{61.62}\\
    Fine-Tuning (T5-base)~\cite{DBLP:journals/jmlr/RaffelSRLNMZLL20}  & \textit{57.61} &\textit{65.39} & \textit{33.96} & \textit{55.53} & \textit{50.50} & \textit{63.59} & \textit{36.49} & \textit{51.96}\\
    Fine-Tuning (T5-large)~\cite{DBLP:journals/jmlr/RaffelSRLNMZLL20}  & \textit{61.37} & \textit{69.69} & \textit{38.04} & \textit{57.65} & \textit{58.12} & \textit{71.13} & \textit{42.04} & \textit{59.40}\\
    \midrule
    Prefix-Tuning (T5-base)~\cite{DBLP:conf/acl/LiL20}  & \textit{25.87} & \textit{52.45} & \textit{7.49} & \textit{31.09}  & \textit{25.08} & \textit{54.18} & \textit{8.45}& \textit{26.37}\\
    Prefix-Tuning (T5-large)~\cite{DBLP:conf/acl/LiL20}  & \textit{63.74} & \textit{71.51} & \textit{39.73} & \textit{59.77} & \textit{58.06} & \textit{69.84} & \textit{36.86} & \textit{56.53}\\
    Prompt-Tuning (T5-base)~\cite{DBLP:conf/emnlp/LesterAC21}  & \textit{30.17} & \textit{56.11} & \textit{15.01} & \textit{38.21} & \textit{25.26} & \textit{55.09} & \textit{8.97 }& \textit{27.68}\\
    Prompt-Tuning (T5-large)~\cite{DBLP:conf/emnlp/LesterAC21}  & \textit{66.95} & \textit{71.99} & \textit{44.08} & \textit{60.15} &  \textit{\underline{59.92}} &  \textit{71.02} & \textit{40.75} & \textit{60.44}\\    
    PCP (RoBERTa-base)~\cite{DBLP:journals/corr/abs-2210-07032}  & 64.95 & 70.84 & 41.55 & 60.54 & \textit{53.00} & \textit{66.58} & \textit{41.19} & \textit{56.14} \\        
    PCP (RoBERTa-large)~\cite{DBLP:journals/corr/abs-2210-07032}  & 67.79 & \underline{73.80} & 44.04 & \underline{61.41}  & \textit{52.75} & \textit{71.13}  & \textit{\underline{43.04}} & \textit{60.44} \\    
    \midrule    
    DiscoPrompt (T5-base) & 65.79 & 71.70 & 43.68 & 61.02  & 64.90 & 71.28  & 41.82 & 59.27\\ 
    DiscoPrompt (T5-large) & \bf70.84 & \bf75.65 & \bf49.03 & \bf64.58 & \bf67.06 & \bf73.76 & \bf45.25 & \bf63.05\\
    \midrule    
    DiscoPrompt (T5-11b) & 75.34 & 78.06 & 52.42 & 68.14  & 72.78 & 77.55 & 47.18 & 67.62\\ 
    \bottomrule
  \end{tabular}}
  \vspace{-0.05in}
  \caption{
  \label{tab:discourse relation classification PDTB 2.0}
The accuracy (\%) and F1 score (\%) are evaluated on the PDTB 2.0 dataset.
\textit{Italics numbers} indicate the results of reproduced models, underlined numbers correspond to the second best. ContrastiveIDRR corresponds to the model without a data augmentation for a fair comparison.
More baselines before 2019 can be found in Table~\ref{tab:Appendix_discourserelation_classification_pdtb2.0} in Appendix.
}
\vspace{-0.5cm}
\end{table*}

\subsection{Dataset}
The experiments are conducted on two datasets, the PDTB 2.0~\cite{DBLP:conf/lrec/PrasadDLMRJW08} and the CoNLL-2016 shared task (CoNLL16)~\cite{DBLP:conf/conll/XueNPRWWW16}, to validate the performance of our method. 
Both contain the Wall Street Journal (WSJ) articles, and the difference is the annotation and relation senses.
We evaluate performance on PDTB 2.0 according to two different settings denoted as \textit{Ji}~\cite{DBLP:journals/tacl/JiE15} and \textit{Lin}~\cite{DBLP:conf/emnlp/LinKN09} with 11 subtypes.
The CoNLL-2016 shared task provides more abundant annotations and two test data denoted as \textit{Test} and \textit{Blind} with 15 subtypes. 
More specific details and statistics are listed in Appendix~\ref{sec:appendix_data_statistics}.

\subsection{Implementation Details}

We employ the T5 model~\cite{DBLP:journals/jmlr/RaffelSRLNMZLL20} as the backbone to implement \textbf{DiscoPrompt} and use the T5-large as the primary model for a fair comparison with extensive baselines.
Generally, the overall configuration follows the setting in \citet{DBLP:conf/emnlp/LesterAC21}, and we put more details of the configuration in Appendix~\ref{sec:appendix_discoPrompt_implementation_details}.
We report the Macro-F1 score and accuracy in experiments and ablation studies.
A prediction is considered as correct whenever it matches one of the ground-truth labels.
All experiments are conducted with 2 $\times$ NVIDIA V100 (32\text{GB}) except for the T5-11b scale on  2 $\times$ NVIDIA A6000 (48\text{GB}).

\subsection{Baselines}
This paper mainly adopts two categories of competitive baselines for the PDTB 2.0 dataset and the CoNLL-2016 shared task\footnote{We report our produced results via the official code if the authors did not report results on those data.}. The first category is the previous state-of-the-art (SOTA) baselines, such as TransS~\cite{DBLP:conf/acl/HeWGH20}, BMGF-RoBERTa~\cite{DBLP:conf/ijcai/LiuOSJ20}, LDSGM~\cite{DBLP:conf/aaai/WuCGLZS22}, XLNet-large~\cite{DBLP:conf/acl/KimFGL20}, OTMT (XLNet-large)~\cite{DBLP:conf/www/JiangQL22}, and ContrastiveIDRR~\cite{conf/emnlp/Contrastive_Learning}.
Two partitions of these SOTA baselines are highlighted for comparison with our method. One partition utilizes the hierarchical information in their methods (e.g., the LDSGM and ContrastiveIDRR), and the other is to fine-tune the pre-trained language models (e.g., XLNet-large). Therefore, we include the fine-tuned T5 models to illustrate the performance gain of prompt tuning. Besides, a prompt-based method PCP~\cite{DBLP:journals/corr/abs-2210-07032} and general Prefix-Tuning~\cite{DBLP:conf/acl/LiL20}, as well as Prompt Tuning~\cite{DBLP:conf/emnlp/LesterAC21} are included. The details of implementation are listed in~\ref{sec:Appendix_baseline_mdoels}.



\section{Experimental Result}

\begin{table*}[t]
\vspace{-0.6cm}
\centering
\small
  \scalebox{0.9}{\begin{tabular}{l| cc | cc | cc | cc }
    \toprule
    \multirow{2}{*}{ \textbf{Models} } &    
    \multicolumn{2}{c|}{\textbf{Test (Top)}} &
    \multicolumn{2}{c|}{\textbf{Test (Second)}} &
    \multicolumn{2}{c|}{\textbf{Blind (Top)}} &
    \multicolumn{2}{c}{\textbf{Blind (Second)}}  \\
    { } & {\textbf{F1}} & {\textbf{Accuracy}} & {\textbf{F1}}& {\textbf{Accuracy}} & {\textbf{F1}} & {\textbf{Accuracy}} & {\textbf{F1}} & {\textbf{Accuracy}} \\
    \midrule
    CoNLL Baseline~\cite{DBLP:conf/conll/RutherfordX16} & - & - & - & 36.13 & - & - & - & 37.67 \\
    MTL-Attn-LSTM~\cite{Lan2017Multi} & - & - & - & 39.40 & - & - & - & 40.12 \\
    RWP-CNN~\cite{DBLP:conf/sigdial/VariaHC19} & - & - & - & 39.39 & - & - & - & 39.36 \\
    BMGF-RoBERTa~\cite{DBLP:conf/ijcai/LiuOSJ20}  & \textit{56.55} & \textit{68.23} & {40.68} & {57.26} & \textit{58.30} & \textit{74.43} & {28.98} & {55.19}  \\    
    
    \midrule
    XLNet (base, cased)~\cite{DBLP:conf/acl/KimFGL20}   &  \textit{43.48} & \textit{62.29} & \textit{18.80} & \textit{33.16} & \textit{19.90} & \textit{66.12} & \textit{9.07 }& \textit{28.71} \\
    XLNet (large, cased)~\cite{DBLP:conf/acl/KimFGL20}  & \textit{47.07} & \textit{64.76} & \textit{27.13} & \textit{47.85} & \textit{22.37} & \textit{66.59} & \textit{11.94} & \textit{35.06} \\
    Fine-Tuning (T5-base)~\cite{DBLP:journals/jmlr/RaffelSRLNMZLL20}  & \textit{54.64} & \textit{67.10} & \textit{31.99} & \textit{53.92}  & \textit{50.94} & \textit{71.30} &\textit{24.52} & \textit{49.89} \\
    Fine-Tuning (T5-large)~\cite{DBLP:journals/jmlr/RaffelSRLNMZLL20}  & \textit{58.74} & \textit{70.87} & \textit{34.66} & \textit{58.88}  & \textit{56.28} &\textit{73.07} & \textit{24.63} & \textit{54.30} \\
    \midrule
    Prefix-Tuning (T5-base)~\cite{DBLP:conf/acl/LiL20}  & \textit{26.18} & \textit{55.35} & \textit{8.26}& \textit{26.63} & \textit{27.17} & \textit{65.88} & \textit{9.70}& \textit{32.71} \\
    Prefix-Tuning (T5-large)~\cite{DBLP:conf/acl/LiL20}  & \textit{57.84} & \textit{71.15} & \textit{46.06} & \textit{59.40} & \textit{55.61} & \textit{74.12} & \textit{30.53} & \textit{55.53}\\
    Prompt-Tuning (T5-base)~\cite{DBLP:conf/emnlp/LesterAC21}  & \textit{25.53} & \textit{54.44} & \textit{13.01} & \textit{29.11} &  \textit{27.21} & \textit{64.71} & \textit{11.55} & \textit{33.65}\\
    Prompt-Tuning (T5-large)~\cite{DBLP:conf/emnlp/LesterAC21}  & \textit{59.95} &\textit{72.32} & \underline{\textit{49.59}} & \underline{\textit{60.57}} & \textit{63.35} & \underline{\textit{77.41}} & \underline{\textit{35.72}} & \underline{\textit{57.88}}\\
    PCP (RoBERTa-base)~\cite{DBLP:journals/corr/abs-2210-07032}  & \textit{58.54} & \textit{69.31} & \textit{33.27} & \textit{55.48} & \textit{55.30} & \textit{72.00}  & \textit{26.00} &\textit{50.99} \\
    PCP (RoBERTa-large)~\cite{DBLP:journals/corr/abs-2210-07032}  & \underline{\textit{63.78}} & \underline{\textit{72.69}} & \textit{37.79 }& \textit{58.36 }& \underline{\textit{64.74}} &\textit{ 76.47}  &\textit{ 27.77} & \textit{56.24 }\\   
    \midrule
    DiscoPrompt (T5-base) & 60.66 & 70.63 & 45.99 & 60.84  & 62.98 &  76.94 &  39.27 & 57.88\\
    DiscoPrompt (T5-large) & \bf 69.56 & \bf 75.33 & \bf 56.29 & \bf 66.32 & \bf 67.89 & \bf 80.47 & \bf 38.49 & \bf 63.06\\
    \midrule
    DiscoPrompt (T5-11b) & 70.38 & 78.07 & 57.75 & 69.71  & 72.33 & 84.94 & 38.60 & 66.35\\ 
    \bottomrule
  \end{tabular}}
  \vspace{-0.05in}
  \caption{
  \label{tab:discourserelation classification CoNLL16}
The accuracy (\%) and F1 score (\%) are evaluated on the implicit discourse partition of CoNLL16 dataset. \textit{Italics number} indicate the results of reproduced models.
}
\vspace{-0.5cm}
\end{table*}

\subsection{Main Results}
Table~\ref{tab:discourse relation classification PDTB 2.0} and Table~\ref{tab:discourserelation classification CoNLL16} summarize the main results of the PDTB 2.0 and CoNLL16 datasets, from which we derive the following conclusions.
\textbf{First}, our method significantly outperforms all baselines and achieves state-of-the-art performance at both top and second-level classes in the IDRR task.
Specifically, our method gains a considerable improvement of 6.93\% second-level accuracy, 10.99\% second-level F1 score, 5.96\% top-level accuracy, and 9.47\% top-level F1 score over the fine-tuning of the T5-large model in PDTB (\textit{Ji}).
It demonstrates that our method effectively utilizes the structure information and perceives the specific knowledge on the correlation of discourse relations and connectives 
and finally enhances the ability of T5 to undertake this challenging task.
\textbf{Second}, the prompt-based baselines (e.g., Prefix-Tuning, Prompt-Tuning, and PCP) receive outstanding performance and perform better than the T5-large fine-tuning method on this task.
Many works~\cite{DBLP:conf/naacl/ScaoR21,DBLP:conf/emnlp/LesterAC21} have discussed the overfitting problem of T5-large fine-tuning, and this can be partially solved by prompt-tuning by updating a few learnable parameters with limited training instances.
The learnable parameters of baselines and DiscoPrompt are shown in Appendix~\ref{sec:appendix_approximation_learnable parameters}.  
\textbf{Third}, the ContrastiveIDRR and our method obtain better F1 scores.
This observation can support the necessity of integrating the dependencies among relations as well as connectives in the label hierarchy.

Fine-tuning a relatively pre-trained large language model (LLM) such as T5-11b requires extensive computation resources to update all trainable parameters.
However, by adapting the prompt tuning-based method, the entire LLM is frozen, and only a few learnable parameters of input embeddings are required to update to obtain satisfactory performance.
Therefore, we also include the performance of DiscoPrompt with the T5-11b version as a reference to explore the ability of a sizeable pre-trained language model on this IDDR task.
As shown in Table~\ref{tab:discourse relation classification PDTB 2.0} and Table~\ref{tab:discourserelation classification CoNLL16}, DiscoPrompt (T5-11b) easily beats other methods, achieving a 52.42\% F1 score and 68.14\% accuracy in the 11-class classification (second-level) task of the PDTB (\textit{Ji}) and illustrating the benefits without adjusting the representations from LLMs.
On the contrary, fine-tuning T5-11b is infeasible in most single compute nodes. Considering the computation cost, we still focus on the comparison among large models.

\begin{table*}[!htbp]
\vspace{-0.6cm}
\small
\centering
\scalebox{0.9}{\begin{tabular}{ll|c|c|c|c}
\toprule
\multicolumn{2}{c|}{\textbf{Model}}
& \textbf{F1 (Top)} & \textbf{Accuracy (Top)} & \textbf{F1 (Second)} & \textbf{Accuracy (Second)} \\
\midrule
\multicolumn{2}{l|}{\quad PCP (RoBERTa-large)~\cite{DBLP:journals/corr/abs-2210-07032}}  & 67.79 & 73.80 & 44.04 & 61.41\\ 
\multicolumn{2}{l|}{\quad DiscoPrompt (T5-large)} & \textbf{70.84} & \textbf{75.65} & \textbf{49.03} & \textbf{64.58} \\
\midrule
   \parbox[t]{10mm}{\multirow{4}{*}{Path}}
   & w/ Top \& Second  & 53.93 & 66.89 & 33.74 & 53.71\\
   & w/ Top \& Connective & 69.19 & 72.57 & 42.95 & 64.08\\
   & w/ Second \& Connective & 70.04 & 74.69 & 45.98 & 64.37\\
   & w/ Connective & 68.00 & 73.82 & 43.76 & 63.43 \\
   & w/ Second & 63.45 & 71.99 & 40.52 & 59.67 \\
\midrule
\midrule
  \parbox[t]{10mm}{\multirow{3}{*}{Prompt}}
  & w/o Entire Discrete Prompt & 68.38 & 72.95 & 41.79 & 62.66 \\ 
  & w/o Cloze Discrete Prompt & 68.64 & 73.72 & 41.44 & 63.72 \\
  & w/o Hierarchical Tree Prompt & 68.03 & 72.18 & 43.14 & 62.85\\
\midrule
\midrule
  \parbox[t]{10mm}{\multirow{3}{*}{Hierarchy}}
  & w/ Continuous Hierarchy Prompt  & 67.63 & 73.24 & 44.03 & 63.81 \\
  & w/ Continuous Labels \& Connective & 67.74  & 73.24 & 44.06 & 64.10 \\
  & w/ Continuous Connective & 68.35  & 73.15 & 44.48 & 64.20 \\  
\bottomrule
\end{tabular}}
\vspace{-0.05in}
\caption{
Ablation study in the components of DiscoPrompt on PDTB (\textit{Ji}).
The path part considers different combinations in the path prediction; the prompt part tries to eliminate templates from the structure-aware prompt; the hierarchy replaces the hierarchical tree prompt with continuous variants.
}
\label{table:methods_ablation}
\vspace{-0.6cm}
\end{table*}

\subsection{Ablation Study}

To better investigate the factors of DiscoPrompt, we design numerous ablations on the path prediction and the tailored hierarchical tree prompt.
Table~\ref{table:methods_ablation} reports the performance of the ablation study for our model in the PDTB (\textit{Ji}).

\paragraph{Joint Probability for Path Prediction}
In our method, by estimating the likelihoods of $p_1^j$ (the top-level), $p_2^j$ (the second-level), and $p_3^j$ (the connective) in a predicted path, the dependencies of these three masks are utilized for enhancing the ability of the pre-trained language model on this IDRR task.
According to the experimental results in Table~\ref{table:methods_ablation}, we can conclude that
1) the performance of the path prediction model incorporating the signals from all three masks surpasses other models (i.e., paths forming by two arbitrary masks or one connective mask), emphasizing the significance of dependencies and effectiveness of joint prediction;
2) the predicted path model without prior knowledge of selected discriminative connectives (i.e., Path w/ Top \& Second) performs the worst, which is consistent with findings in~\citet{DBLP:journals/corr/abs-2210-07032};
3) the predicted path model with only the connective mask (e.g., Path w/ Connective) performs consistently worse than paths adding the second mask, indicating the slight ambiguity of connectives and the necessity of the label hierarchy especially with the top.
The performance gain with the complete path is at least 3.76\% on average, and models associating with paths including individual connective masks can also  beat the previous SOTA.



\paragraph{Discrete Prompt Template} Two portions in our designed prompt template are in natural textual form and as discrete non-tunable tokens.
The first part is the discrete tokens for the label hierarchy structure (i.e., \textbf{hierarchical tree prompt}), shown in Figure~\ref{fig:DiscoursePrompt_Architecture} and Figure~\ref{fig:DiscoPrompt_Template_Searching}.
The second part is the \textbf{cloze discrete prompt} ``The path is''.
We remove the discrete tokens from the template to evaluate their importance.
The performance shown in Table~\ref{table:methods_ablation} demonstrates that the two parts of the prompt are essential for achieving satisfactory performance compared with the without manual tips (i.e., Prompt w/o Entire Prompt).
When adding back the cloze discrete prompt, we do not observe the model's ability to understand the correlations among masks for path prediction.
Without explicitly injecting structural information into the hierarchical tree prompt, the performance dropped significantly, especially the second-level F1 score, dropping from 49.03\% to 43.14\%.


\begin{table}[t]
\small
\centering
\setlength\tabcolsep{2pt}
\scalebox{0.85}{\begin{tabular}{l|c|c|c|c}
\toprule
\multicolumn{1}{c|}{\textbf{Model}}
& \textbf{Comp.} & \textbf{Cont.} & \textbf{Exp.} & \textbf{Temp.} \\
\midrule
MTL-MLoss\scriptsize{~\cite{DBLP:conf/acl/NguyenLTN19}} & 48.44 & 56.84 & 73.66 & 38.60 \\
KANN\scriptsize{~\cite{DBLP:conf/aaai/GuoHDW20}} & 43.92 & 57.67 & 73.45 & 36.33 \\
BMGF-RoBERTa\scriptsize{~\cite{DBLP:conf/ijcai/LiuOSJ20}} & 59.44 & 60.98 & 77.66 & 50.26 \\
CG-T5\scriptsize{~\cite{DBLP:conf/emnlp/JiangFCLZ21}} & 55.40 & 57.04 & 74.76 & 41.54 \\
CVAE\scriptsize{~\cite{DBLP:conf/emnlp/DouHSZ21}} & 55.72 & 63.39 & 80.34 & 44.01\\
ContrastiveIDRR\scriptsize{~\cite{conf/emnlp/Contrastive_Learning}} & 65.84 & 63.55 & 79.17 & \underline{69.86}\\
\hline
DiscoPrompt (T5-base) & 62.55  & 64.45  & 78.77 & 57.41 \\
DiscoPrompt (T5-large) & \underline{67.13}  & \underline{69.76} & \underline{81.61} & 64.86 \\
\hline
DiscoPrompt (T5-11b) & \bf74.35  & \bf 72.44 & \bf82.57 & \bf72.00 \\
\bottomrule
\end{tabular}}
\vspace{-0.05in}
\caption{The performance for top-level classes on PDTB (\textit{Ji}) in terms of F1 (\%) (top-level multi-class classification). More baselines for comparison can be found in Table~\ref{table:appendix_binary} in Appendix~\ref{sec:appendix_Performance on Top and Second level}.}
\label{table:binary}
\vspace{-0.6cm}
\end{table}


\paragraph{Hierarchical Tree Prompt}
To acquire a deeper understanding of the discrete hierarchical tree prompt, we perform experiments to gradually replace the discrete tokens with continuous ones in various elements of this hierarchy prompt. The experiments include 1) Continuous Hierarchy Prompt: replacing the whole hierarchical tree prompt as the continuous tunable prompt with the same number of tokens, 2) Continuous Labels \& Connective: only including the ``-> ''  and replacing other relation labels and connective as continuous tunable prompt, and 3) Continuous Connective: only replacing the textual connective to be the tunable prompt. The experimental result in Table~\ref{table:methods_ablation} underscores the importance and effectiveness of our tailored discrete hierarchical tree prompt, which obtains at least 4.98\% performance boost.


\paragraph{Prompt Engineering} Furthermore, we conduct the prompt template searching and the parameter sensitivity on the continuous prompt length that we describe in Appendix~\ref{sec: Appendix_ablation_study_DiscoPrompt}.

\begin{table}[t]
\small
\centering
\setlength\tabcolsep{2pt}
\scalebox{0.85}{\begin{tabular}{l|c|c|c|c}
\toprule
\textbf{Second-level Label}  & \textbf{PCP} & \textbf{Contrast}  & \textbf{DP (large)} & \textbf{DP (11b)} \\
\midrule
\textit {Temp.Asynchronous} & 57.81 & 59.79  & \underline{64.15} & \bf72.27\\
\textit {Temp.Synchrony} & 0.0 & \bf78.26 & \underline{50.00} & 33.33\\
\hline
\textit {Cont.Cause} &  65.64 & 65.58  & \underline{69.66} & \bf72.28\\
\textit {Cont.PragmaticCause} & 0.0 & 0.0 & 0.0 & 0.0\\
\hline
\textit {Comp.Contrast}  & \underline{63.88} & 62.63 &  62.88 & \bf70.63\\
\textit {Comp.Concession}  & \underline{8.00} & 0.0 & \bf 9.09 & 0.0 \\
\hline
\textit {Exp.Conjunction} & 57.78 & 58.35 &  \underline{60.09} & \bf62.84\\
\textit {Exp.Instantiation} & 74.01 & 73.04  & \underline{74.17} & \bf76.60\\
\textit {Exp.Restatement} & 61.00 & 60.00 & \underline{65.24} & \bf 65.98\\
\textit {Exp.Alternative} & \underline{66.67} & 53.85  &  60.00 & \bf84.21\\
\textit {Exp.List} & 29.63 & \underline{34.78} & 24.00 & \bf38.46\\
\bottomrule
\end{tabular}}
\vspace{-0.05in}
\caption{The label-wise F1 scores for the second-level labels on PDTB (\textit{Ji}) (second-level multi-class classification).
``Contrast'' and ``DP'' indicate the ContrastiveIDRR and DiscoPrompt.
Results of more baselines are listed in Table~\ref{table:Appendix_Label_wise_F1} in Appendix~\ref{sec:appendix_Performance on Top and Second level}.}
\vspace{-0.6cm}
\label{table:Label-wise F1}
\end{table}

\subsection{Label-wise F1 Scores}
The PDTB (\textit{Ji}) setting exhibits highly skewed label distributions, with only roughly 854 training instances (i.e., 6.8\% of 12406 training instances) annotating as five of the 11 second-level labels. To further explore our model in four top-level relations and 11 second-level sense types on this dataset, Table~\ref{table:binary} and Table~\ref{table:Label-wise F1} report the F1 scores (\%) of the top-level and second-level classes, respectively.
In Table~\ref{table:binary}, our model outperforms all baselines in three top-level relations (i.e., \textit{Comparison}, \textit{Contingency}, \textit{Expansion}), and most of the baselines in the \textit{Temporal} relation except ContrastiveIDRR.
Specifically, Table~\ref{table:Label-wise F1} illustrates that our model performs better on the \textit{Temp.Asynchronous} second-level class, whereas ContrastiveIDRR is much better on the \textit{Temp.Synchrony}.
In Table~\ref{table:Label-wise F1}, our model obtains valid predictions on most second-level classes, but all methods fail to predict \textit{Cont.Pragmatic Cause}. 
This situation may result from the few training examples of this class being insufficient for optimal learnable parameters, and the models tend to ignore this class in the prediction process.
When we check the less representative classes (i.e., \textit{Temp.Synchrony}, \textit{Comp.Concession}), DiscoPrompt can still make correct predictions, while PCP and ContrastiveIDRR still fail to predict neither correct ones. Moreover, we can also see the power of LLMs that the T5-11b performs remarkably better than smaller models.

\subsection{Prompt Adaptation} 
\begin{figure}[t]
    \vspace{-0.9cm}
    \centering
    \includegraphics[width=0.95\linewidth]{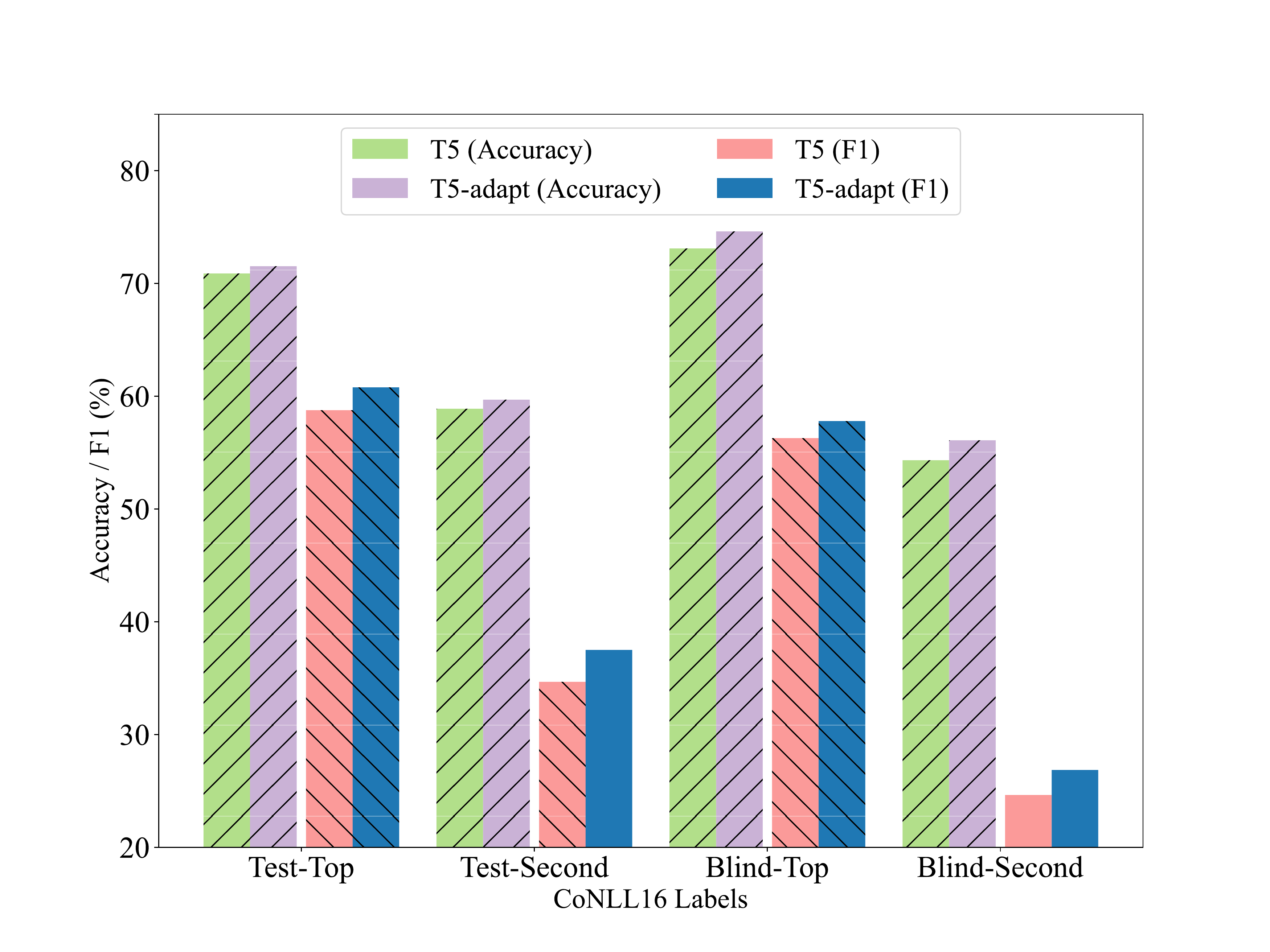}
    \vspace{-0.1in}
    \caption{The performance comparison of the T5-large fine-tuning with and without using our designed template on the CoNLL16 dataset.
    }
    \label{fig:T5_finetune_template_CoNLL2016}
    \vspace{-0.6cm}
\end{figure}

For T5 Fine-Tuning \label{sec:Designed Prompt For T5 Fine-Tuning}
To demonstrate the effectiveness of our designed template and explore whether our designed template can be used for the fine-tuning paradigm, we convert the data input to the tailored prompt template
but with only a [MASK] for generating the entire path.
The experimental results on CoNLL16 are summarised in Figure~\ref{fig:T5_finetune_template_CoNLL2016}, and the T5-adapt boosts all metrics over vanilla T5-large fine-tuning.
The detailed performance and the experimental results for PDTB 2.0 are shown in Table~\ref{table:T5-large-fine-tune-template-table} and Figure~\ref{fig:T5_finetune_template_PDTB2.0} in Appendix~\ref{appendix_Performance of Designed Prompt For T5 Fine-Tune}.

\begin{table}[!t]
\vspace{-0.9cm}
\small
\centering
\setlength\tabcolsep{2pt}
\scalebox{0.85}{
\begin{tabular}{l|c|c|c|c}
\toprule
\multicolumn{1}{c|}{\textbf{Model}}
& \textbf{F1 (Top)} & \textbf{Acc (Top)} & \textbf{F1 (Second)} & \textbf{Acc (Second)} \\
\hline
Random & 23.44 & 32.18 & 6.48 & 8.78 \\
ChatGPT$_\text{label}$ & 43.37 & 48.51 & 16.17 & 26.95  \\
ChatGPT$_\text{label \& con.}$ & 43.99 & 49.28 & 17.55& 30.32\\
ChatGPT$_\text{structure}$ & 44.09 & 50.24 & 19.88 & 31.95 \\
\hline
\end{tabular}
}
\vspace{-0.4cm}
\caption{
The performance of ChatGPT performs on the PDTB (\textit{Ji}) test set. ChatGPT$_{label \& con.}$ means predicting the label and connective, and ChatGPT$_{structure}$ means adopting our structural path prompt template.
}
\label{tab:ChatGPT_Performance}
\vspace{-0.3cm}
\end{table}

\subsection{Prompt Adaptation For ChatGPT}
With the powerful ability of LLMs exhibited on numerous tasks, we are curious about the capability of ChatGPT on zero-shot IDRR task.
We test the ability of ChatGPT with three designed templates on the PDTB (\textit{Ji}), and the performance is shown in Table~\ref{tab:ChatGPT_Performance}.
All designed templates obtain higher performance than the random, but still at a low region in the second level compared with supervised learning.
This result reveals that IDRR is still tricky for ChatGPT and cannot solve easily at current state, consistent with the result in \citet{DBLP:journals/corr/abs-2304-14827}.
The structural path template outperforms the other two templates, proving the help of the structural form for ChatGPT to understand this task.
The F1 score of each second level is shown in Figure~\ref{fig:appendix_ChatGPT_performance_figure} in Appendix and illustrates the effectiveness to distinguish various second-level senses among the \textit{Expansion} top class.
More case examples and discussions refer to Appendix~\ref{sec:appendix_Discussion for ChatGPT and Case Example}.

\begin{table}[!t]
\small
\centering
\setlength\tabcolsep{2pt}
\begin{tabular}{l|c|c}
\hline
\textbf{Model} & \textbf{Acc} & \textbf{F1}\\
\hline
\citet{DBLP:conf/acl/PitlerN09} & 94.15 & - \\ 
\citet{DBLP:conf/naacl/DaiH18}  & 94.46 & 93.70 \\
\citet{Dai2019A}                           & 95.39 & 94.84\\
\citet{DBLP:journals/corr/abs-2210-07032}      & 94.78 & 93.59\\
\citet{DBLP:conf/sigdial/VariaHC19}          & 96.20 & 95.48\\ 
\hline 
Fine-tuning (T5-large) w/o Connective & 74.47 & 72.38\\ 
Fine-tuning (T5-large) w/ Gold Connective & 95.41 & 94.94\\ 
DiscoPrompt (T5-large) w/ Connective Mask & 78.35 & 74.62\\
DiscoPrompt (T5-large) w/ Gold Connective  & \textbf{96.73} & \textbf{95.64}\\ 
\hline
\end{tabular}
\vspace{-0.1in}
\caption{
Explicit Top-level sense classification results on PDTB (\textit{Ji}).
``w/o Connective'' and ``w/ Connective Mask'' regard the EDRR as IDRR.
}
\label{tab:EDRR_Performance}
\vspace{-0.7cm}
\end{table}

\subsection{Generalization to Explicit Discourse Relation Classification Task}\label{sec:generalizationEDRR}
To demonstrate the generalization ability of our model, we transfer and adapt our method to the explicit discourse relation recognition (EDRR) task.
We simply replace the first [MASK] between two arguments with the gold connective for each instance in EDRR.
Following the previous works~\cite{DBLP:conf/sigdial/VariaHC19, DBLP:journals/corr/abs-2210-07032}, the second-level class is the same as our implicit one setting.
In Table~\ref{tab:EDRR_Performance}, our model slightly outperforms previous SOTA models on the top-level sense prediction. DiscoPrompt consistently outperforms fine-tuning under different settings,  
and we observe a larger margin with absenting connectives.

\section{Conclusion}
In this paper, we introduce a path prediction method for tackling the IDRR task by utilizing the hierarchical structural information and prior knowledge of connectives.
Combining label structures in natural language with prompt tuning successfully takes a step further in this task as well as other generalized settings, e.g., prompt adaptation and explicit relation detection.
Our model achieves new SOTA performance on PDTB 2.0 and CoNLL-2016 data, and we hope our detailed discussions can help communities in discourse fields. 
\section*{Limitations and Future Work}
\paragraph{Limited Utilized Knowledge} The main limitation of our method is the limited utilized knowledge. Since our prompt tuning-based method tests on Implicit Discourse Relation Recognition (IDRR) task, the elicited knowledge only comes from the dataset of this task and the model pre-training corpora. This constraint restricts the capability owing to the reporting bias~\cite{DBLP:conf/cikm/GordonD13} in the pre-training models (PLMs). Moreover, the relatively few training data of several second-level classes resulting from the highly skewed label distribution problem requires extensive knowledge to make the model understand data instances and the task. Although we impose the prior human knowledge against the IDRR task from the input template designing to the discourse connectives selection, the knowledge source still only comes from our prior knowledge and the elicited knowledge of PLMs. As a result, even our method obtains a valid score in all second-level classes except the~\textit{Cont.Pragmatic Cause} displayed in Table~\ref{table:Label-wise F1}, some second-level senses, which are the same as previous studies, cannot receive a satisfactory performance (e.g.,~\textit{Comp.Concession} and~\textit{Expa.List}). The future work for this issue is to integrate more abundant knowledge and equip the model with more vital abilities. For example, grounding the arguments pair on the relevant nodes of the knowledge graph for each data instance~\cite{DBLP:conf/emnlp/LinCCR19} or knowledge distillation from large language models to provides more contextual information and enhances the capability of the model on this task.

\paragraph{Limited Predicted Connectives} Another area for improvement is the prediction of extensive connectives. Although our model includes the pre-selected connectives as our third layer of a designed hierarchy tree, we do not include the ground truth of connectives as our third layer. Because including these extensive connectives to form many leaves will result in many paths (more than 100). This limitation may be addressed in future works by utilizing the pruning algorithms for reducing a lot of redundant nodes and leaves on each instance to enhance effectiveness and efficiency.

\section*{Ethics Statement}
In this work, we conformed to recognized privacy practices and rigorously followed the data usage policy. We declare that all authors of this paper acknowledge the \emph{ACM Code of Ethics} and honor the code of conduct. This paper presents a method to utilize the interaction information between different layers, inherent sense label structure, and prior knowledge of connectives in the implicit discourse recognition task. The PDTB 2.0 and  CoNLL-2016 dataset were used to train and assess the ability of the pre-trained language model on this task. The PDTB2.0 and CoNLL2016-Test dataset is collected from the Wall Street Journal (WSJ) articles, while the CoNLL2016-Blind dataset is derived from newswire texts, the primary language is English based and belongs to the news domain. We can foresee no immediate social consequences or ethical issues as we do not introduce social/ethical bias into the model or amplify any bias from the data. Therefore, these two datasets are not required to perform further actions to check the offensive content.

\section*{Acknowledgements}
The authors of this paper were supported by the NSFC Fund (U20B2053) from the NSFC of China, the RIF (R6020-19 and R6021-20) and the GRF (16211520 and 16205322) from RGC of Hong Kong, the MHKJFS (MHP/001/19) from ITC of Hong Kong and the National Key R\&D Program of China (2019YFE0198200) with special thanks to HKMAAC and CUSBLT. We also thank the support from NVIDIA AI Technology Center (NVAITC) and the UGC Research Matching Grants (RMGS20EG01-D, RMGS20CR11, RMGS20CR12, RMGS20EG19, RMGS20EG21, RMGS23CR05, RMGS23EG08).

\bibliography{anthology}
\bibliographystyle{acl_natbib}
\newpage
\appendix

\section{Appendix for Experimental Settings}\label{sec:appendix}

\subsection{DataSet} \label{sec:appendix_data_statistics}
\paragraph{The Penn Discourse Treebank 2.0 (PDTB 2.0)}
PDTB 2.0\footnotemark[2] is a large-scale corpus containing 2,312 Wall Street Journal (WSJ) articles~\cite{DBLP:conf/lrec/PrasadDLMRJW08}, that employs a lexically-grounded approach to annotating discourse relations. This corpus includes three sense levels (i.e., classes, types, and sub-types) and naturally forms the sense hierarchy. In this dataset, we validate our model on two popular settings of the PDTB 2.0 dataset, which are the Ji-setting~\cite{DBLP:journals/tacl/JiE15} and Lin-setting~\cite{DBLP:conf/emnlp/LinKN09}. The former one following~\citet{DBLP:journals/tacl/JiE15} to split sections 2-20, 0-1, and 21-22 as training, validation, and test sets respectively, while the latter follows ~\citet{DBLP:conf/emnlp/LinKN09} split sections 2-21, 22, 23 as training, validation, and test sets respectively. We evaluate our model on the four top-level implicit discourse relations and the 11 major second-level implicit discourse senses by following previous works~\cite{DBLP:conf/aaai/WuCGLZS22, conf/emnlp/Contrastive_Learning,DBLP:journals/corr/abs-2210-07032}. The data statistics of the top-level and second-level senses are displayed in Table~\ref{table:pdtb_statistics} and Table~\ref{table:pdtb_Sec_level_statistics}.

\paragraph{The CoNLL-2016 Shared Task (CoNLL16)}
The CoNLL-2016 shared task\footnotemark[3]~\cite{DBLP:conf/conll/XueNPRWWW16} provides more abundant annotation (e.g., second-level sense type) for shadow discourse parsing. This task includes two test sets, the PDTB section 23 (CoNLL-Test) and newswire texts (CoNLL-Blind), that comply with the PDTB annotation guidelines. Compared with PDTB 2.0, CoNLL16 includes more new class sense (e.g., \textit{Contingency.Condition}) and merges several labels to annotate new labels. For example, \textit{Contingency.Pragmatic cause} is
merged into \textit{Contingency.Cause.Reason} to remove the former type with very few samples. In this paper, we follow~\citet{DBLP:conf/conll/WangL16, Lan2017Multi, DBLP:conf/ijcai/LiuOSJ20} to perform the experiments on this CoNLL-2016 dataset and validate the performance of our model in the top- and second-level sense.  

\footnotetext[3]{The License of the PDTB 2.0 dataset is LDC User Agreement for Non-Members, and this paper is consistent with their intended use for research purposes. This dataset download from~\url{https://catalog.ldc.upenn.edu/LDC2008T05}}
\footnotetext[4]{ CoNLL16 dataset download from~\url{https://www.cs.brandeis.edu/~clp/conll16st/dataset.html}.}

\begin{table}[t]
\small
\centering
\begin{tabular}{l|c|c|c}
\toprule
\textbf{Top-level Senses}  & \textbf{Train} & \textbf{Val.} & \textbf{Test} \\
\midrule
Comparison & 1,942 & 197 & 152 \\
Contingency & 3,342 & 295 & 279 \\
Expansion & 7,004 & 671 & 574 \\
Temporal & 760 & 64 & 85 \\
\hline
Total & 12,362 & 1,183 & 1,046\\
\bottomrule
\end{tabular}
\caption{Statistics of four top-level implicit senses in PDTB 2.0.}
\label{table:pdtb_statistics}
\end{table}

\begin{table}[t]
\small
\centering
\begin{tabular}{l|c|c|c}
\toprule
\textbf{Second-level Senses}  & \textbf{Train} & \textbf{Val.} & \textbf{Test} \\
\midrule
Comp.Concession & 180 & 15 & 17\\
Comp.Contrast & 1566 & 166 & 128\\
Cont.Cause & 3227 & 281 & 269\\
Cont.Pragmatic cause & 51 & 6 & 7\\
Exp.Alternative & 146 & 10 & 9\\
Exp.Conjunction & 2805 & 258 & 200\\
Exp.Instantiation & 1061 & 106 & 118\\
Exp.List & 330 & 9 & 12\\
Exp.Restatement & 2376 & 260 & 211\\
Temp.Asynchronous & 517 & 46 & 54\\
Temp.Synchrony & 147 & 8 & 14\\
\hline
Total &  12406 & 1165 & 1039\\
\bottomrule
\end{tabular}
\caption{The implicit discourse relation data statistics of second-level types in PDTB 2.0.}
\label{table:pdtb_Sec_level_statistics}
\end{table}

\subsection{DiscoPrompt Implementation Details} \label{sec:appendix_discoPrompt_implementation_details}
DiscoPrompt is prompt tuning upon T5-model, and we also validate our method over various model scales, including T5-base, T5-large, and T5-11b. Figure~\ref{tab:conll2016_label} shows the heat map of highly frequent connectives on CoNLL2016, and the label words are in Table~\ref{tab:conll2016_label}. Generally, the overall configuration follows the setting in \citet{DBLP:conf/emnlp/LesterAC21} and sets the learnable prompt length as 20. The training was implemented using cross-entropy loss with 30,000 training steps, which selects the model that yields the best performance on the validation set. We adopt an Adafactor~\cite{DBLP:conf/icml/ShazeerS18} optimizer with various learning rate ranges for different dataset settings. The batch size and maximum input sequence are 4 and 350, respectively. The maximum generates sequence length of the encoder is 10. Our model is conducted on two 32GB NVIDIA V100 GPUs, except for the T5-11b scale on two 48GB NVIDIA A6000 GPUs. The running time for T5-base is around 8 hours, while T5-large is about 19 hours. 

Since we are interested in the ability of our method to adopt a larger-scale model on this task, we tested the T5-11b model on various datasets. Most of the configuration is the same as the above T5-large version. The slight differences in hyperparameters are batch size is one and gradient accumulation step is 16. The running time of the T5-11b model is around 50 hours. The tailored prompt template is shown in Figure~\ref{fig:DiscoPrompt_Template_Searching}. The specific hyperparameters of implementation details for DiscoPrompt (T5-large) and DiscoPrompt (T5-11b) are displayed in Table~\ref{table:Implementation_details_hypreparameters}. The frozen pre-train T5 model download from \textit{HuggingFace}, and our model inheritance and modification from \textit{OpenPrompt}~\cite{DBLP:conf/acl/DingHZCLZS22}.

\begin{table}[t]
\small
\centering
\begin{tabular}{l|c}
\toprule
\textbf{Dataset} & \textbf{Hyperparameters} \\

\midrule
\multirow{2}{*}{PDTB (\textit{Ji})}     & LR space: \{9e-2, 9e-1\}, LR$^{*}$: 3e-1,\\
                                & BS: 4, gradient accumulation step:1 \\
\midrule
\multirow{2}{*}{PDTB (\textit{Lin})}    & LR space: \{9e-4, 9e-3\}, LR$^{*}$: 2e-4,\\
                                & BS: 4, gradient accumulation step:1\\
\midrule                               
\multirow{2}{*}{CoNLL16 (\textit{Test})}   & LR space: \{9e-2, 9e-1\}, LR$^{*}$: 9e-2,\\
                                & BS: 4, gradient accumulation step:1\\
\midrule
\multirow{2}{*}{CoNLL16 (\textit{Blind})}  & LR space: \{9e-2, 9e-1\}, LR$^{*}$: 9e-2,\\
                                & BS: 4, gradient accumulation step:1\\
\midrule
\midrule
\multirow{2}{*}{PDTB (\textit{Ji})}     & LR space: \{9e-4, 9e-3\}, LR$^{*}$: 4e-4,\\
                                & BS: 1, gradient accumulation step:16 \\
\midrule
\multirow{2}{*}{PDTB (\textit{Lin})}    & LR space: \{9e-4, 9e-3\}, LR$^{*}$: 5e-4,\\
                                & BS: 1, gradient accumulation step:16\\
\midrule                               
\multirow{2}{*}{CoNLL16 (\textit{Test})}   & LR space: \{9e-5, 9e-4\}, LR$^{*}$: 9e-5\\
                                & BS: 1, gradient accumulation step:16\\
\midrule
\multirow{2}{*}{CoNLL16 (\textit{Blind})}  & LR space: \{9e-4, 9e-3\}, LR$^{*}$: 2e-4,\\
                                & BS: 1, gradient accumulation step:16\\
\bottomrule 
\end{tabular}
\caption{The hyperparameters of implementation details for DiscoPrompt (T5-large) and DiscoPrompt (T5-11b). The upper part is for the T5-large version in four datasets, while the bottom is for the T5-11b version. ``LR space'', ``LR$^{*}$'', and ``BS'' refer to learning rate searching space, optimal learning rate, and batch size, respectively. }
\label{table:Implementation_details_hypreparameters}
\end{table}

\begin{table}[t]
\centering
\small
  \scalebox{0.8}{\begin{tabular}{c| c | c}
  \toprule
  \textbf{Top-level} &  \textbf{Second-level} &  \textbf{Connectives}\\
  \midrule
  \multirow{2}{*}{Comparison} & Concession & nonetheless \\
                              & Contrast & but\\
  \midrule
  \multirow{3}{*}{Contingency} & Reason & because \\
                               & Result & so \\
                               & Condition & if\\
  \midrule
  \multirow{6}{*}{Expansion} & Alternative & unless \\
                             & Chosen & instead \\
                             & Conjunction & and \\ 
                             & Instantiation & for example \\ 
                             & Exception & except \\ 
                             & Restatement & indeed\\                              
  \midrule
  \multirow{3}{*}{Temporal} & Precedence & before \\
                            & Succession & previously \\
                            & Synchrony & when \\
  \bottomrule
  \end{tabular}}
  \caption{
  \label{tab:conll2016_label}
The label word set on CoNLL2016 dataset.
}
\vspace{-0.6cm}
\end{table}

\begin{figure}[t]
    \vspace{-0.25in}
    \centering
    \includegraphics[width=\linewidth]{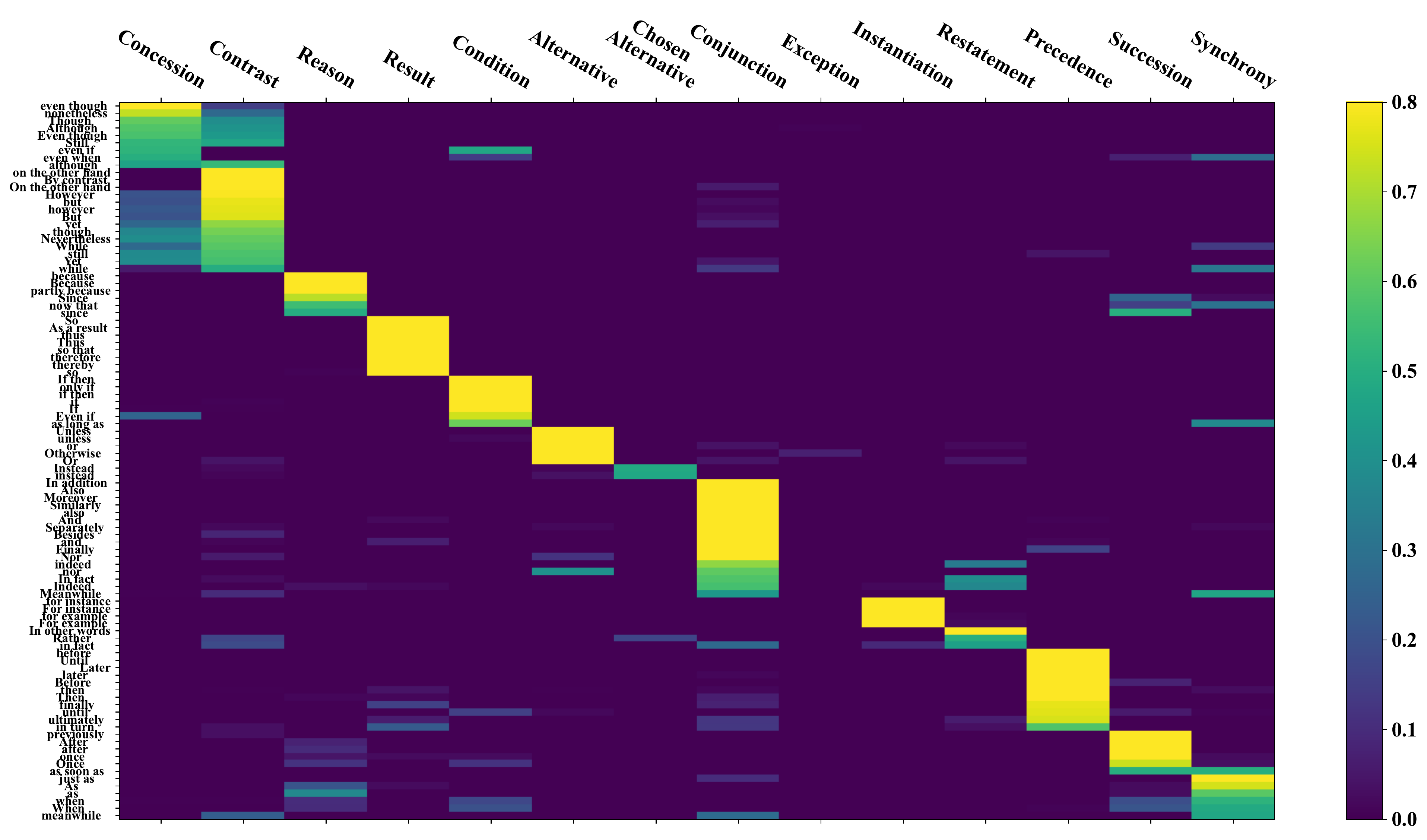}
    \vspace{-0.3in}
    \caption{Prior probabilities of CoNLL16 frequent connectives.
    }
    \label{fig:conll2016_heatmap}
    \vspace{-0.2in}
\end{figure}


\subsection{Baseline Models} \label{sec:Appendix_baseline_mdoels}
To exhibit the effectiveness of our proposed method, we compared it with previous works on the PDTB 2.0 and CoNLL16 datasets. In this section, we mainly describe some recently published baselines, and more baselines can be found in Table~\ref{tab:Appendix_discourserelation_classification_pdtb2.0}.

\paragraph{Common Baselines for PDTB 2.0 and CoNLL16:} 
\begin{itemize}[leftmargin=*]
    \item \textbf{RWP-CNN}~\cite{DBLP:conf/sigdial/VariaHC19}: a convolutional neural networks-based method to model word pairs in the arguments in a discourse relation.
    
    \item \textbf{BMGF-RoBERTa}~\cite{DBLP:conf/ijcai/LiuOSJ20}: a RoBERTa-based model, which contains a robust contextualized representation module, a bilateral matching module to capture the interaction between arguments, and a global information fusion module to derive final representations for labels.

    \item \textbf{XLNet}~\cite{DBLP:conf/acl/KimFGL20}: it fine-tunes XLNet model~\cite{DBLP:conf/nips/YangDYCSL19} for IDRR task to predict the flat label in each layer of discourse relation sense.
    
    \item \textbf{T5 (Fine-Tuning)}~\cite{DBLP:journals/jmlr/RaffelSRLNMZLL20}: Fine-tune a T5-model based on specifics tailored input text in various settings with a comparison of our model. The Implementation details are described in Appendix~\ref{sec: Appendix_t5_finetune_implementation}. 
    
    \item \textbf{Prefix-Tuning (T5)}~\cite{DBLP:conf/acl/LiL20}: a lightweight method concatenates the tunable prefix tokens before the discrete input text, keeps language model parameters frozen, and optimizes these continuous task-specific prefix tokens. The implementation details of the Prefix-Tuning methods are appended in Appendix~\ref{sec:implement_details_prompt_baseline}.
    
    \item \textbf{Prompt-Tuning (T5)}~\cite{DBLP:conf/emnlp/LesterAC21}: a vanilla Prompt Tuning-based model conditioning on a frozen model, releasing the constraints of the prompt templates from discrete to learnable prompts. The implementation details of the prompt tuning methods are appended in Appendix~\ref{sec:implement_details_prompt_baseline}.
    
    \item \textbf{PCP}~\cite{DBLP:journals/corr/abs-2210-07032}: a prompt-based connective prediction method for IDRR by adopting the RoBERTa model. This method utilizes the strong correlation between connectives and discourse relations to map the predicted connectives to respective implicit discourse relations.

\end{itemize}

\paragraph{Baselines for PDTB 2.0:}
\begin{itemize}[leftmargin=*]
    \item \textbf{DER}~\cite{Bai2018Deep}: a model enhanced with multiple grained text representations, including character, subword, word, sentence, and sentence pair levels. 
    
    \item \textbf{MTL-MLoss}~\cite{DBLP:conf/acl/NguyenLTN19}: a multi-task learning neural model that predicts labels and connectives simultaneously by leveraging the dependence between them.
    
    \item \textbf{ELMo-C\&E}~\cite{Dai2019A}: a neural model that employs a regularization approach to utilize the external event knowledge and coreference relations.
    
    \item \textbf{TransS}~\cite{DBLP:conf/acl/HeWGH20}: a TransS-driven joint learning model which translates the discourse relations in low-dimensional embedding space (i.e., TransS), and simultaneously learns the semantic features of arguments.
    
    \item \textbf{CG-T5}~\cite{DBLP:conf/emnlp/JiangFCLZ21}: a joint model that recognizes the relation label and generates the desired target sentence containing the meaning of relations simultaneously.
    
    \item \textbf{OTMT(XLNet)}~\cite{DBLP:conf/www/JiangQL22}: an XLNet~\cite{DBLP:conf/nips/YangDYCSL19} based model exploits the knowledge distillation (KD) technique for discourse relation recognition task.
    
    \item \textbf{LDSGM}~\cite{DBLP:conf/aaai/WuCGLZS22}: a label dependence-aware sequence generation model that integrates the global representation of an input instance,  level-specific contexts, and the label dependence decoded by graph convolutional network (GCN) to obtain better label embeddings, and then employ the label sequence decoder to output the predicted labels.
    \item \textbf{GOLF}~\cite{DBLP:journals/corr/abs-2211-13873}: a global and local hierarchy-aware contrastive framework, to model and capture the information from these two kinds of hierarchies with the aid of contrastive learning.
    
    \item \textbf{ContrastiveIDRR}~\cite{conf/emnlp/Contrastive_Learning}: a contrastive learning method for incorporating the sense hierarchy into the recognition process and using the hierarchy to select the negative examples.

\end{itemize}

\paragraph{Baselines for CoNLL16:}
\begin{itemize}[leftmargin=*]

    \item \textbf{CoNLL Baseline}~\cite{DBLP:conf/conll/RutherfordX16}: a neural classifier requires word vectors and a simple feed-forward training procedure. 

    \item \textbf{MTL-Attn-LSTM}~\cite{Lan2017Multi}: a multi-task attention-based LSTM neural network model that exploits explicit discourse relations in PDTB and unannotated external data in a multi-task joint learning framework.
\end{itemize}

\subsection{Implementation Details of the Prefix-Tuning and Prompt Tuning} \label{sec:implement_details_prompt_baseline}
In our paper, we implement the prefix tuning~\cite{DBLP:conf/acl/LiL20} and prompt tuning~\cite{DBLP:conf/emnlp/LesterAC21} methods as the baselines for comparison with our model. We proposed several templates for searching for their best performance in these two methods. The experimental details for these two methods include the template and hyperparameter search. Moreover, there are 154 tokens, including textual tokens (non-tunable tokens) and tunable tokens, in our prompt template. For a fair comparison, we insert 154 tunable tokens into the respective prompt template in these two baselines.

\begin{figure*}[!t]
    \centering
    \includegraphics[width=\linewidth]{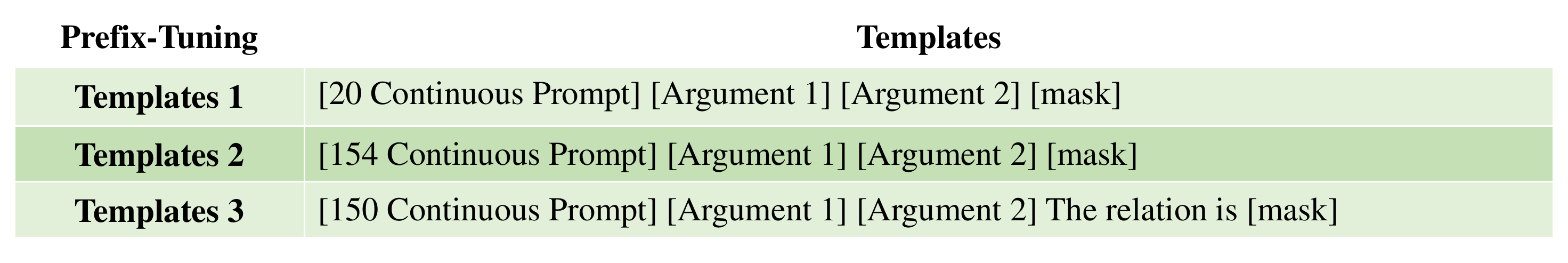}
    \vspace{-0.8cm}
    \caption{Prefix-Tuning Template Searching}
    \label{fig:Prefix_prompt}
\end{figure*}

\paragraph{Prefix-Tuning} Following the setting of prefix tuning~\cite{DBLP:conf/acl/LiL20}, we implemented several designed templates on the PDTB 2.0 JI setting and the templates shown in figure~\ref{fig:Prefix_prompt}. In these templates, we find that the \textbf{prefix-prompt template three} is better among all templates, and we adopted this template for further comparison with our method. The overall configuration of this model follows the settings of prefix tuning~\cite{DBLP:conf/acl/LiL20}. The batch size and maximum sequence length of this model are 8 and 350. The training is performed using cross-entropy loss with an Adafactor optimizer~\cite{DBLP:conf/icml/ShazeerS18} and a learning rate selecting in {0.3, 0.5, 0.8} yields the best performance on the validation set, and the training steps are 30,000. 

\begin{figure*}[!t]
    \vspace{-0.5 cm}
    \centering
    \includegraphics[width=\linewidth]{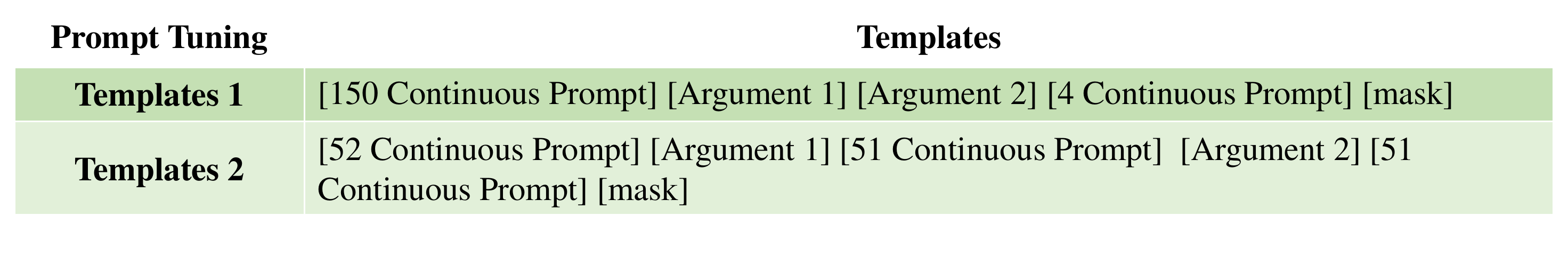}
    \vspace{-1.0 cm}
    \caption{Prompt Tuning Template Searching}
    \label{fig:Prompt_Tuning}
\end{figure*}

\paragraph{Prompt-Tuning} For the prompt tuning method, we implemented several designed templates on the PDTB 2.0 JI setting and the templates shown in figure~\ref{fig:Prompt_Tuning}. In these templates, we find that the \textbf{prompt tuning template two} is better among all templates, and adopted this template for further comparison with our method. The overall configuration of this model follows the settings of prefix tuning~\cite{DBLP:conf/emnlp/LesterAC21}. The batch size and maximum sequence length of this model are 8 and 350. The training is performed using cross-entropy loss with an Adafactor optimizer~\cite{DBLP:conf/icml/ShazeerS18} and a learning rate selecting in {0.3, 0.5, 0.8} yields the best performance on the validation set, and the training steps are 30,000.





\subsection{Implementation Details of T5 Model Fine-Tuning} \label{sec: Appendix_t5_finetune_implementation}
Here we provide the fine-tuning details for T5 base and large models on various datasets.

\noindent \textbf{Model Input and Output}
In main experiments, T5-model fine-tuning as the competitive baseline, we concatenate two arguments with an ``</s>'' at the end of the sequence as input. The T5 model asked to generate the top-level labels, and the second-level labels with concatenating by commas (e.g., Comparison.Contrast) given the data input. For the experiments to test the transferred template on the fine-tuning paradigm, the ``T5-adapt'' model in section~\ref{sec:Designed Prompt For T5 Fine-Tuning} concatenate the hierarchy tree prompt in Figure~\ref{fig:DiscoPrompt_Template_Searching} before the two arguments as input. Then we concatenate a prompt message ``The path is '' before the original output. Furthermore, for the setting ``T5-large (fine-tune) (w/ connective)'' in the EDRR task (Section~\ref{sec:generalizationEDRR}), it required inserting the connectives between two arguments. Therefore, we use the text span named ``FullRawText'' in the dataset with an additional ``</s>'' at the end as input.

\noindent \textbf{Hyperparameter Search}
We first conduct a preliminary experiment to determine the range of hyper-parameters. Then, we search for the learning rate within $\{3e-4, 1e-4\}$ and warmup steps within $\{0, 100\}$. For the T5-base model, we set the training batch size as 8, and the model is evaluated with a batch size of 128 every 150 steps. For the T5-large model, the training and evaluation batch sizes are set as 16 and 64, respectively. The model is optimized with an AdamW optimizer with a linear learning rate schedule. The test performance of the model with the best validation accuracy is reported.

\subsection{The Approximation of Learnable Parameters} \label{sec:appendix_approximation_learnable parameters}
To show the efficiency of our method, we append the approximation of learnable parameters for all models, including our model and baselines. The approximation of learnable parameters is listed in Table~\ref{tab:tunable_parameters}. 

\begin{table}[!t]
\small
\centering
\scalebox{0.9}{\begin{tabular}{l|c }
\toprule
\textbf{Model} &  \textbf{Parameters}\\
\hline
BMGF-RoBERTa~\cite{DBLP:conf/ijcai/LiuOSJ20}                    & 2.3M \\ 
XLNet(base, cased)~\cite{DBLP:conf/acl/KimFGL20}                & 110M \\
XLNet(large, cased)~\cite{DBLP:conf/acl/KimFGL20}               & 340M \\
OTMT(XLNet-base)~\cite{DBLP:conf/www/JiangQL22}                 & 110M \\
OTMT(XLNet-large)~\cite{DBLP:conf/www/JiangQL22}                & 340M \\ 
Fine-Tuning (T5-base)~\cite{DBLP:journals/jmlr/RaffelSRLNMZLL20}   & 220M \\
Fine-Tuning (T5-large)~\cite{DBLP:journals/jmlr/RaffelSRLNMZLL20}  & 770M \\
Prefix-Tuning (T5-base)~\cite{DBLP:conf/acl/LiL20}               & 0.12M \\ 
Prefix-Tuning (T5-large)~\cite{DBLP:conf/acl/LiL20}              & 0.16M \\  
Prompt-Tuning (T5-base)~\cite{DBLP:conf/emnlp/LesterAC21}         & 0.12M  \\ 
Prompt-Tuning (T5-large)~\cite{DBLP:conf/emnlp/LesterAC21}        & 0.16M \\ 
LDSGM~\cite{DBLP:conf/aaai/WuCGLZS22}                           & 128M  \\
ContrastiveIDRR~\cite{conf/emnlp/Contrastive_Learning}          & 125M \\
PCP(RoBERTa-base)~\cite{DBLP:journals/corr/abs-2210-07032}      & 124M \\     
PCP(RoBERTa-large)~\cite{DBLP:journals/corr/abs-2210-07032}     & 335M \\  
\hline
DiscoPrompt (T5-base)                                           & 1.2M \\ 
DiscoPrompt (T5-large)                                          & 2.1M \\ 
\bottomrule
\end{tabular}}
\caption{The approximation of learnable parameters for models. ``M'' stands for million learnable parameters. 
}
\label{tab:tunable_parameters}
\end{table}

\section{Appendix for Evaluation Result and Analysis}\label{sec:evaluation_result_analysis}
\subsection{Performance of Baselines in PDTB 2.0}\label{sec:appendix_Baselines for PDTB 2.0}
In this section, we list extensive baselines in Table~\ref{tab:Appendix_discourserelation_classification_pdtb2.0} for comparison with our method.  

\begin{table*}[!ht]
\centering
\small
  \begin{tabular}{l| cc | cc | cc | cc }
    \toprule
    \multirow{2}{*}{\textbf{Models} } &    
    \multicolumn{2}{c|}{\textbf{Ji (Top)}} &
    \multicolumn{2}{c|}{\textbf{Ji (Sec)}} &
    \multicolumn{2}{c|}{\textbf{Lin (Top)}} &
    \multicolumn{2}{c}{\textbf{Lin (Sec)}} \\
    & {\textbf{F1}} & {\textbf{Acc}} & {\textbf{F1}}& {\textbf{Acc}} & {\textbf{F1}} & {\textbf{Acc}} & {\textbf{F1}} & {\textbf{Acc}} \\
    \midrule
    \citet{DBLP:conf/emnlp/LinKN09} & - & - & -  & - & - & - & - & 40.20 \\
    \citet{DBLP:journals/tacl/JiE15} & - & - & -  & 44.59 & - & - & - & - \\
    \citet{Liu2016Implicit} & 44.98 & 57.27 & - & - & - & - & - & - \\
    \citet{DBLP:conf/coling/QinZZ16}& - & - & - & 45.04 & - & - & - & 43.81 \\
    \citet{Liu2016Recognizing} & 46.29 & 57.57 & - & - & - & - & - & - \\
    \citet{DBLP:conf/acl/WuSCSW17}& 44.84 & 58.85  & - & - & - & - & - & - \\ 
    \citet{Lan2017Multi} & 47.80 & 57.39 & - & - & - & - & - & - \\
    \citet{DBLP:conf/acl/QinZZHX17}& - & - & - & 46.23 & - & - & - & 44.65 \\ 
    \citet{Xu2018Using} & 44.48 & 60.63 & - & - & -  & - & - & -\\
    \citet{DBLP:conf/naacl/DaiH18} & 48.82 &  57.44  & - & - & - & - & - & - \\ 
    \citet{Bai2018Deep} & 51.06 & - & - & 48.22  & - & - & - & 45.73 \\
    \citet{DBLP:conf/iwcs/ShiD19}& 46.40 & 61.42 & - & 47.83 & - & - & - & 45.82 \\  
    \citet{DBLP:conf/sigdial/VariaHC19} & 50.20 & 59.13 & - & - & - & - & - & -\\ 
    \citet{Dai2019A} & 52.89 & 59.66 & - & 48.23  & - & - & - & -\\
    \citet{DBLP:conf/acl/NguyenLTN19} & 53.00 & - & - & 49.95& - & - & - & 46.48\\
    \citet{Shi2019Next} & - & - & - & 53.23 & - & - & - & - \\
    \citet{DBLP:conf/acl/HeWGH20} & - & - & - & - & 51.24 & 59.94 & - & -\\ 
    \citet{DBLP:conf/aaai/GuoHDW20}& 47.90 & 57.25 & - & - & - & - & - & -\\ 
    \citet{DBLP:conf/lrec/KishimotoMK20}& 58.48 & 65.26 & - & 54.32 & - & - & - & -\\
    \citet{DBLP:conf/ijcai/LiuOSJ20}  & 63.39&  69.06  & 35.25 & 58.13 & 58.54 & 68.66 & 39.15 & 53.96 \\
    \citet{DBLP:conf/emnlp/JiangFCLZ21} & 57.18 & - & 37.76 & - & - & - & - & -\\
    \citet{DBLP:journals/corr/abs-2106-03192} & 59.24 &  -  & 39.33 & 55.42 & - & - & - & - \\
    \citet{DBLP:conf/emnlp/DouHSZ21} & 65.06 & 70.17 & - & - & - & - & - & -\\
    \citet{DBLP:conf/aaai/WuCGLZS22} & 63.73 &  71.18 & 40.49 & 60.33 & - & - & - & - \\
    \citet{DBLP:journals/corr/abs-2211-13873} & 65.76 & 72.52 & 41.74 & 61.16 & - & - & - & -  \\
    \citet{conf/emnlp/Contrastive_Learning}(w/o data augm.)\footnotemark[4] & 67.85 &  71.70 & 45.54 & 59.19 & - & - & - & - \\
    \citet{conf/emnlp/Contrastive_Learning} (w data augm.) & 69.60 &  72.18 & 49.66 & 61.69 & - & - & - & - \\    
    \midrule
    BERT-base~\cite{DBLP:conf/naacl/DevlinCLT19} & 43.17 & 62.14 & 26.32 & 50.24 & 43.44 & 63.46 & 26.70 & 49.87\\ 
    BERT-large~\cite{DBLP:conf/naacl/DevlinCLT19} & 57.06 & 67.59 & 30.02 & 54.57 & 56.06 & 68.40 & 38.68 & 56.53\\  
    
    XLNet(base, cased)~\cite{DBLP:conf/acl/KimFGL20} & 59.33 & 66.35 & 36.36 & 54.73 & 56.16 & 68.05 & 36.23 & 55.82\\
    XLNet (large, cased)~\cite{DBLP:conf/acl/KimFGL20}  & 63.58 & 69.52 & 38.24 & 61.29 & 58.97 & 72.17 & 40.71 & 58.77\\
    OTMT (XLNet-base)~\cite{DBLP:conf/www/JiangQL22} &  60.78 & 68.89 & - & 56.65 & - & - & - & 56.37\\
    OTMT (XLNet-large)~\cite{DBLP:conf/www/JiangQL22} & 64.46 & 72.34 & - & 61.06 & - & - & - & 61.62\\
    Fine-Tuning (T5-base)~\cite{DBLP:journals/jmlr/RaffelSRLNMZLL20} & 57.61 & 65.39 & 33.96 & 55.53 & 50.50 & 63.59 & 36.49 & 51.96\\
    Fine-Tuning (T5-large)~\cite{DBLP:journals/jmlr/RaffelSRLNMZLL20} & 61.37 & 69.69 & 38.04 & 57.65 & 58.12 & 71.13 & 42.04 & 59.40\\
    \midrule
    Prefix-Tuning (T5-base)~\cite{DBLP:conf/acl/LiL20} & 25.87 & 52.45 & 7.49  & 31.09  & 25.08 & 54.18 & 8.45 & 26.37\\
    Prefix-Tuning (T5-large)~\cite{DBLP:conf/acl/LiL20} & 63.74 & 71.51 & 39.73 & 59.77 & 58.06 & 69.84 & 36.86 & 56.53\\
    Prompt-Tuning (T5-base)~\cite{DBLP:conf/emnlp/LesterAC21} & 30.17 & 56.11 & 15.01 & 38.21 & 25.26 & 55.09 & 8.97 & 27.68\\
    Prompt-Tuning (T5-large)~\cite{DBLP:conf/emnlp/LesterAC21} & 66.95 & 71.99 & 44.08 & 60.15 &  59.92 &  71.02 & 40.75 & 60.44\\
    PCP w/ RoBERTa-base~\cite{DBLP:journals/corr/abs-2210-07032} & 64.95 & 70.84 & 41.55 & 60.54 & 53.00 & 66.58 & 41.19 & 56.14 \\     
    PCP w/ RoBERTa-large~\cite{DBLP:journals/corr/abs-2210-07032} & 67.79 & 73.80 & 44.04 & 61.41  & 52.75 & 71.13  & 43.04 & 60.44 \\    
    \midrule    
    DiscoPrompt (T5-base)& 65.79 & 71.70 & 43.68 & 61.02  & 64.90 & 71.28  & 41.82 & 59.27\\ 
    DiscoPrompt (T5-large)& \bf70.84 & \bf75.65 & \bf49.03 & \bf64.58 & \bf67.06 & \bf73.76 & \bf45.25 & \bf63.05\\
    \midrule    
    DiscoPrompt (T5-11b)& 75.34 & 78.06 & 52.42 & 68.14  & 72.78 & 77.55 & 47.18 & 67.62\\ 
    \bottomrule
  \end{tabular}
  \caption{
  \label{tab:Appendix_discourserelation_classification_pdtb2.0}
The accuracy (\%) and F1 score (\%) are evaluated on the PDTB 2.0 dataset.}
\vspace{-0.05in}
\end{table*}
\footnotetext[5]{We use their model without a data augmentation version for a fair comparison in Table~\ref{tab:discourse relation classification PDTB 2.0}. This model with the data augmentation version is also appended in this table.}

\subsection{Ablation study on the DiscoPrompt} \label{sec: Appendix_ablation_study_DiscoPrompt}
\begin{figure*}[t]
    \vspace{-0.5 cm}
    \centering
    \includegraphics[width=\linewidth]{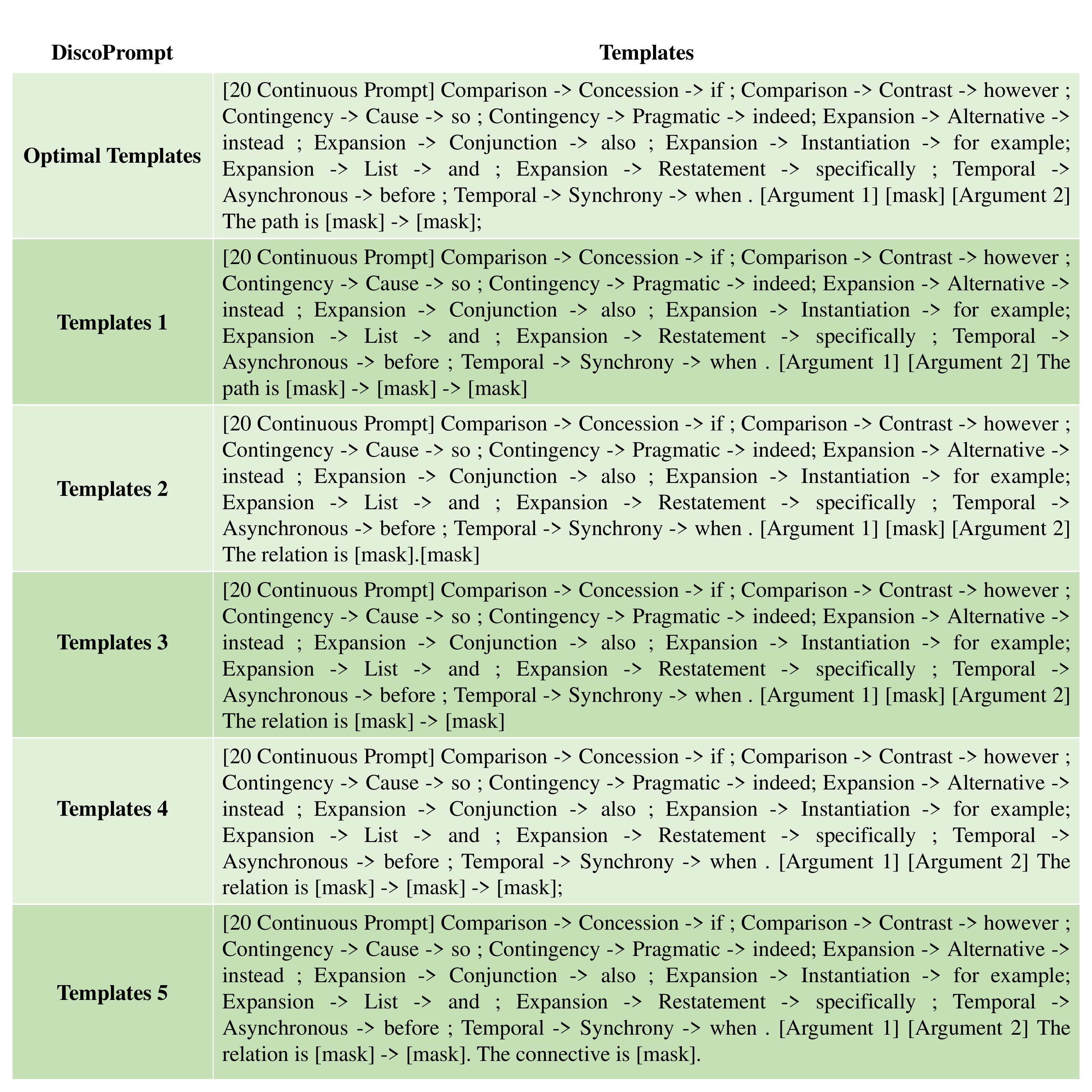}
    \vspace{-1.0 cm}
    \caption{DiscoPrompt Template Searching. The ``Optimal Templates'' is the finalized optimal template for implementing experiments to compare with extensive baselines.}
    \label{fig:DiscoPrompt_Template_Searching}
\end{figure*}

\paragraph{Prompt Template Searching}
We perform the prompt template research on our designed prompt, and all prompt searching templates are listed in Figure ~\ref{fig:DiscoPrompt_Template_Searching}, and the performance is shown in Table ~\ref{table:Appendix_Prompt_Template_Searching}. Our finalized optimal template inserts the connectives between two arguments to improve the textual coherence of input context and results in the PLMs easy to understand input. Therefore,  this template performs better than other designed templates. 

\paragraph{Continuous Prompt Length}
The continuous prompt (i.e., learnable prompt tokens) length is another factor that influences the performance of our model. Hence, we implement various prompt lengths of 10, 20, 50, and 100. The performance is in Table~\ref{table:appendix_continuous_length_ablation}, and the optimal continuous prompt length is 20, which provides the best performance among all the prompt lengths and is the default prompt length for implementing other experiments. Adopting more prompt length than 20 on our method will not significantly increase this task's performance on various evaluation metrics.  

\begin{table*}[t]
\small
\centering
\scalebox{0.85}{\begin{tabular}{l|c|c|c|c}
\toprule
\multicolumn{1}{c|}{\textbf{Model}}
& \textbf{F1 (Top)} & \textbf{Acc (Top)} & \textbf{F1 (Second)} & \textbf{Acc (Second)} \\
\midrule
DiscoPrompt (Optimal Template : The path is {mask} -> {mask};) & \textbf{70.84} & \textbf{75.65} & \textbf{49.03} & \textbf{64.58} \\
\midrule
DiscoPrompt (Template 1: The path is {mask} -> {mask} -> {mask}) & 69.22 & 73.44 & 43.52 & 63.33 \\
DiscoPrompt (Template 2: The relation is {mask}.{mask}) & 67.55  & \underline{74.21} & 44.81 & \underline{64.20}  \\
DiscoPrompt (Template 3: The relation is {mask} -> {mask})& \underline{69.70} & 74.01 & \underline{48.61} & 64.10\\
DiscoPrompt (Template 4: The relation is {mask} -> {mask} -> {mask};) & 68.07 & 72.76 & 45.91 & 62.56 \\
DiscoPrompt (Template 5: The relation is {mask} -> {mask}.The connective is {mask} .) & 62.71 & 70.74 & 40.19 & 58.81 \\
\bottomrule
\end{tabular}}
\caption{Performance of various templates of our method with adopting T5-large model in PDTB (\textit{Ji}) dataset. The details of various templates are shown in Figure~\ref{fig:DiscoPrompt_Template_Searching}.}
\label{table:Appendix_Prompt_Template_Searching}
\end{table*}

\begin{table*}[!htbp]
\small
\centering
\scalebox{0.95}{\begin{tabular}{l|c|c|c|c}
\toprule
\multicolumn{1}{c|}{\textbf{Model}}
& \textbf{F1 (Top)} & \textbf{Acc (Top)} & \textbf{F1 (Second)} & \textbf{Acc (Second)} \\
\midrule
DiscoPrompt (T5-large) & \textbf{70.84} & \textbf{75.65} & \textbf{49.03} & \textbf{64.58} \\
\midrule
   Continuous Prompt Length (10) & 67.17 & 72.47 & 43.56 & 62.66\\
   Continuous Prompt Length (50) & 69.64 & 74.40 & 45.06 & 63.91\\
   Continuous Prompt Length (100) & 68.39 & 73.92 & 42.77 & 64.20\\
\bottomrule
\end{tabular}}
\vspace{-0.05in}
\caption{
Performance of various continuous prompt lengths in our method DiscoPrompt (T5-large) on PDTB (\textit{Ji}) dataset. The default continuous prompt length of our model is 20. 
}
\label{table:appendix_continuous_length_ablation}
\vspace{-0.3cm}
\end{table*}

\subsection{Performance of Label-wise F1 Score on Top and Second level }\label{sec:appendix_Performance on Top and Second level}
The performance (F1 score\%) of more baselines for comparison with our model in Top-level and Second-level shown in Table~\ref{table:appendix_binary} and Table~\ref{table:Appendix_Label_wise_F1}.

\begin{table}[t]
\small
\centering
\setlength\tabcolsep{2pt}
\scalebox{0.9}{\begin{tabular}{l|c|c|c|c}
\toprule
\multicolumn{1}{c|}{\textbf{Model}}
& \textbf{Comp.} & \textbf{Cont.} & \textbf{Exp.} & \textbf{Temp.} \\
\midrule
\citet{DBLP:journals/tacl/JiE15} & 35.93 & 52.78 & - & 27.63 \\
\citet{DBLP:conf/naacl/RutherfordX15} & 41.00 & 53.80 & 69.40 & 33.30 \\
\citet{Liu2016Implicit}  & 37.91 & 55.88 & 69.97 & 37.17 \\
\citet{Liu2016Recognizing}\footnotemark[2] & 39.86 & 54.48 & 70.43 & 38.84 \\
\citet{Qin2016A} & 38.67 & 54.91 & 71.50 & 32.76 \\
\citet{Lan2017Multi}  & 40.73 & 58.96 & 72.47 & 38.50 \\
\citet{Bai2018Deep} & 47.85 & 54.47 & 70.60 & 36.87 \\
\citet{DBLP:conf/naacl/DaiH18} & 46.79 & 57.09 & 70.41 & 45.61 \\
\citet{DBLP:conf/sigdial/VariaHC19}  & 44.10 & 56.02 & 72.11 & 44.41 \\
\citet{DBLP:conf/acl/NguyenLTN19} & 48.44 & 56.84 & 73.66 & 38.60\\
\citet{DBLP:conf/aaai/GuoHDW20} & 43.92 & 57.67 & 73.45 & 36.33\\
\citet{DBLP:conf/ijcai/LiuOSJ20} & 59.44 & 60.98 & 77.66 & 50.26 \\
\citet{DBLP:conf/emnlp/JiangFCLZ21} & 55.40 & 57.04 & 74.76 & 41.54\\
\citet{DBLP:conf/emnlp/DouHSZ21} & 55.72 & 63.39 & 80.34 & 44.01\\
\citet{conf/emnlp/Contrastive_Learning} & 65.84 & 63.55 & 79.17 & \underline{69.86}\\
\hline
DiscoPrompt (T5-base)& 62.55  & 64.45  & 78.77 & 57.41 \\
DiscoPrompt (T5-large)& \underline{67.13}  & \underline{69.76} & \underline{81.61} & 64.86 \\
\hline
DiscoPrompt (T5-11b)& \bf74.35  & \bf 72.44 & \bf82.57 & \bf72.00 \\
\bottomrule
\end{tabular}}
\caption{The performance for top-level classes on PDTB (\textit{Ji}) in terms of F1 (\%) (top-level multi-class classification).}
\label{table:appendix_binary}
\end{table}

\begin{table}[t]
\small
\centering
\scalebox{0.7}{\begin{tabular}{l|c|c|c|c}
\toprule
\multicolumn{1}{c|}{Model(DataSet Settings)}
 & Acc (Second)  & F1 (Second) & Acc (Top) &  F1 (Top)\\
\midrule
   T5 (PDTB (\textit{Ji}))        & 57.65 & 38.04 & 69.69 & 61.37\\   
   T5-adapt(PDTB (\textit{Ji})) & 59.77 & 38.08 & 70.17 & 60.89\\
\midrule 
   T5 (PDTB (\textit{Lin}))        & 59.40 & 42.04 & 71.13 & 58.12\\   
   T5-adapt(PDTB (\textit{Lin})) & 59.53 & 42.83 & 71.91 & 61.03\\
\midrule 
    T5 (CoNLL-Test)        & 58.88 & 34.66&70.87&58.74\\
    T5-adapt(CoNLL-Test) & 59.66 & 37.49 & 71.52 & 60.78\\
\midrule 
    T5 (CoNLL-Blind)        &54.3 & 24.63 & 73.07 & 56.28\\
    T5-adapt(CoNLL-Blind) & 56.07 & 26.85 & 74.61 & 57.77\\
\bottomrule
\end{tabular}}
\caption{The performance comparison of the T5-large fine-tuning with and without using our designed template on the PDTB 2.0 and CoNLL16 dataset. ``T5-adapt'' means adopting our designed template in the fine-tuning process. Acc and F1 inside the brackets indicate the accuracy and F1 score. }
\label{table:T5-large-fine-tune-template-table}
\end{table}

\begin{table*}[b]
\small
\centering
\scalebox{0.9}{\begin{tabular}{l|c|c|c|c|c|c|c}
\toprule
\textbf{Second-level Label} & \textbf{BMGF} & \textbf{LDSGM} & \textbf{PCP} & \textbf{ContrastiveIDRR} & $\textbf{Ours}_{(\textbf{base})}$ & $\textbf{Ours}_{(\textbf{large})}$ & $\textbf{Ours}_{(\textbf{11B})}$ \\

\midrule
\textit {Temp.Asynchronous} & 56.18 & 56.47 & 57.81 & 59.79 & 57.69 & 64.15 & \bf72.27\\
\textit {Temp.Synchrony} & 0.0 & 0.0 & 0.0 & \bf78.26 & 0.0 & 50.00 & 33.33\\
\hline
\textit {Cont.Cause} & 59.60 & 64.36 & 65.64 & 65.58 & 63.83  &  69.66 & \bf72.28\\
\textit {Cont.PragmaticCause} & 0.0 & 0.0 & 0.0 & 0.0 & 0.0 & 0.0 & 0.0\\
\hline
\textit {Comp.Contrast} & 59.75 & 63.52 & 63.88 & 62.63 & 59.26  &  62.88 & \bf70.63\\
\textit {Comp.Concession} & 0.0 & 0.0 & 8.00 & 0.0 & \bf 9.09 & \bf 9.09 & 0.0 \\
\hline
\textit {Expa.Conjunction} & 60.17 & 57.91 & 57.78 & 58.35 &  61.08 &  60.09 & \bf62.84\\
\textit {Expa.Instantiation} & 67.96 & 72.60 & 74.01 & 73.04 & 69.96 & 74.17 & \bf76.60\\
\textit {Expa.Restatement} & 53.83 & 58.06 & 61.00 & 60.00 & 58.45 & 65.24 & \bf 65.98\\
\textit {Expa.Alternative} & 60.00 & 63.46 & 66.67 & 53.85 & 72.73 &  60.00 & \bf84.21\\
\textit {Expa.List} & 0.0 & 8.98 & 29.63 & 34.78 &  37.50 & 24.00 & \bf38.46\\
\bottomrule
\end{tabular}}
\vspace{-0.05in}
\caption{The label-wise F1 scores for the second-level labels on PDTB (\textit{Ji}) (second-level multi-class classification).}
\label{table:Appendix_Label_wise_F1}
\end{table*}

\subsection{Performance of Designed Prompt For T5 Fine-Tuning} \label{appendix_Performance of Designed Prompt For T5 Fine-Tune}
\begin{figure}[!htbp]
    \centering
    \includegraphics[width=0.95\linewidth]{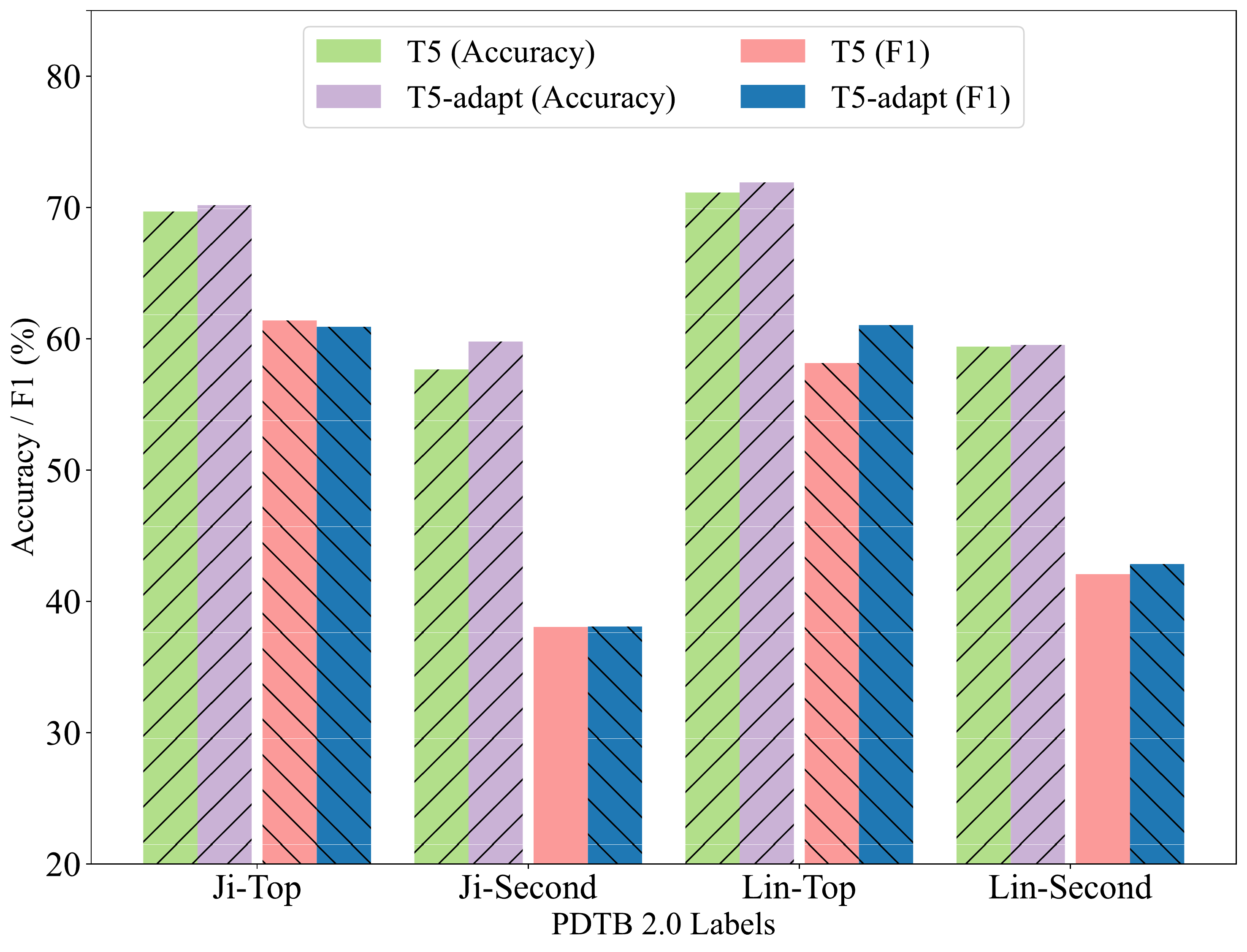}
    \vspace{-0.1in}
    \caption{The performance comparison of the T5-large fine-tuning with and without using our designed template on the PDTB 2.0 dataset. ``T5-adapt'' means adopting our designed template in the fine-tuning process. Acc and F1 inside the brackets indicate the accuracy and F1 score.  
    }
    \label{fig:T5_finetune_template_PDTB2.0}
    \vspace{-0.8cm}
\end{figure}

The performance comparison of the T5-large fine-tuning with and without using our designed template on the PDTB 2.0 is displayed in Figure~\ref{fig:T5_finetune_template_PDTB2.0}. The detailed experimental result for PDTB 2.0 and CoNLL16 dataset is shown in Table~\ref{table:T5-large-fine-tune-template-table}.

\subsection{Discussion and Case Example for ChatGPT}\label{sec:appendix_Discussion for ChatGPT and Case Example}
\begin{figure*}[!ht]
    \centering
    \includegraphics[width=0.6\linewidth]{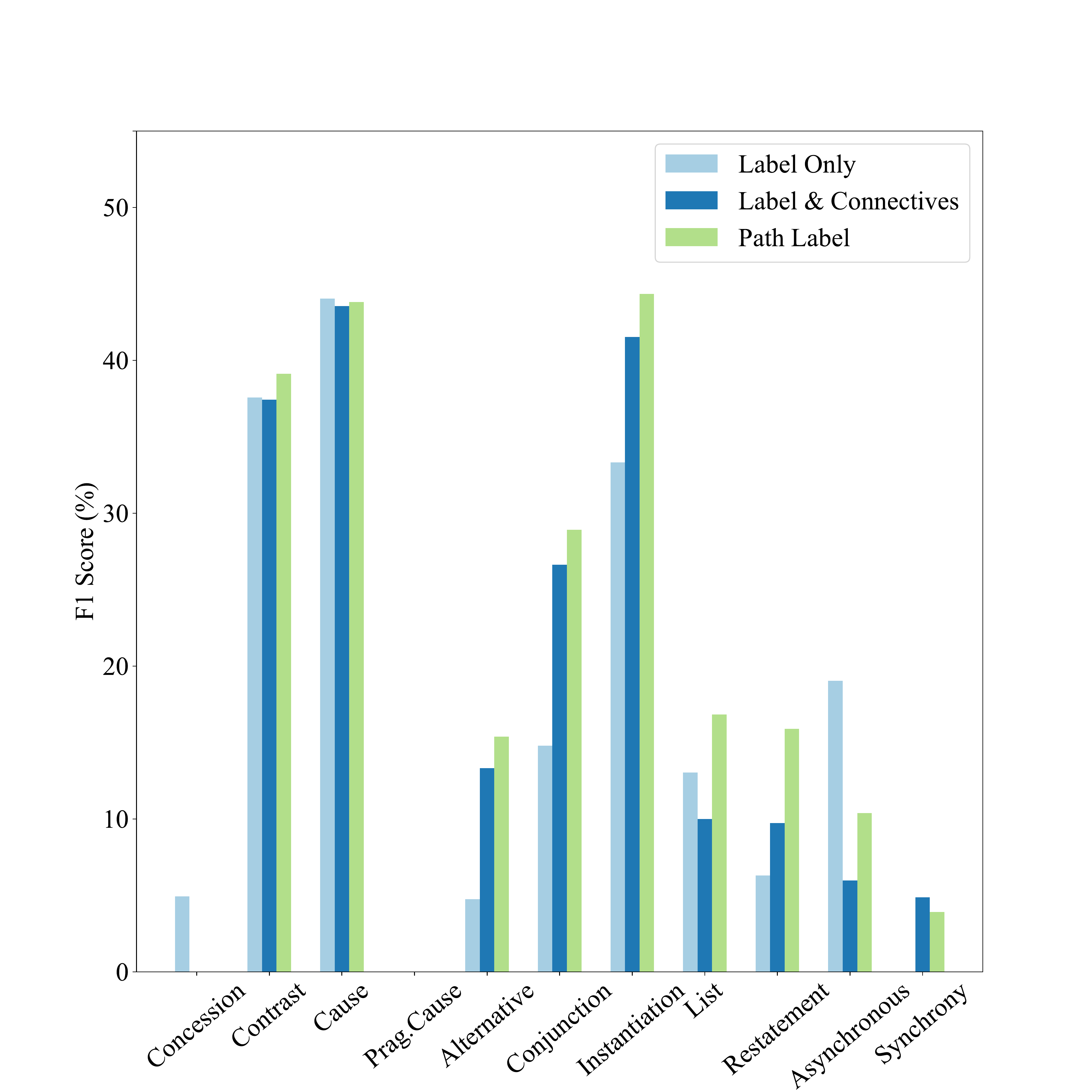}
    \vspace{0.1in}
    \caption{The performance comparison of various input prompt templates for ChatGPT. ``Prag.Cause'' stand for \textit{Pragmatic cause} second level sense.   
    }
    \label{fig:appendix_ChatGPT_performance_figure}
\end{figure*}

We test the ability of ChatGPT~\footnotemark[5] with three designed templates on the PDTB (\textit{Ji}) test set. These templates include: 1) predict the class label only, 2) predict the class label with connectives, and 3) predict the class label with connectives in a structural path form. Moreover, the input template with in-context learning highly relies on the training examples selected as the prefix instruction part of the prompt template. The performance of this model is high variance with the chosen examples vary. Therefore, this template is not taken into account in this section. The performance of the random guess model is obtained by averaging the performance of 5 runs. A prediction is regarded as wrong if ChatGPT generates the answer out of the range of label words. An interesting finding is that the ChatGPT with label-only template tends to predict many temporally related instances to the \textit{Contingency.Cause} second-level sense result in poor performance on \textit{Temporal.synchrony} second-level sense shown in Figure ~\ref{fig:appendix_ChatGPT_performance_figure}. The input template and two case examples are shown in Table~\ref{table:appendix_ChatGPT_Case_Study_1} and Table~\ref{table:appendix_ChatGPT_Case_Study_2}.

\footnotetext[6]{The demonstration and details of ChatGPT are on the website \url{https://openai.com/blog/chatgpt/}}

\begin{table}[!htbp]
\small
\centering
\begin{tabular}{m{24em}}
\toprule
\textbf{Query Template:} Argument 1: Right away you notice the
following things about a Philip Glass concert. Argument 2:
It attracts people with funny hair. What is the relation
label between Argument 1 and Argument 2? Select from
the candidates.\\
1. Comparison.Concession\\
2. Comparison.Contrast\\
3. Contingency.Cause\\
4. Contingency.Pragmatic\\
5. Expansion.Alternative\\
6. Expansion.Conjunction\\
7. Expansion.Instantiation\\
8. Expansion.List\\
9. Expansion.Restatement\\
10. Temporal.Asynchronous\\
11. Temporal.Synchrony\\
\textbf{ChatGPT:} Expansion.Instantiation\\
\midrule
\midrule
\textbf{Query Template:} Argument 1: Right away you notice the
following things about a Philip Glass concert. Argument 2:
It attracts people with funny hair. What is the relation
and connective between Argument 1 and Argument 2?
Select from the candidates.\\
1. Comparison.Concession, if\\
2. Comparison.Contrast, however\\
3. Contingency.Cause, so\\
4. Contingency.Pragmatic, indeed\\
5. Expansion.Alternative, instead\\
6. Expansion.Conjunction, also\\
7. Expansion.Instantiation, for example\\
8. Expansion.List, and\\
9. Expansion.Restatement, specifically\\
10. Temporal.Asynchronous, before\\
11. Temporal.Synchrony, when\\
\textbf{ChatGPT:} Expansion.Instantiation, for example\\
\midrule
\midrule
\textbf{Query Template:} Argument 1: Right away you notice the
following things about a Philip Glass concert. Argument 2:
It attracts people with funny hair. What is the relation
path between Argument 1 and Argument 2? Select from
the candidates.\\
1. Comparison -> Concession -> if\\
2. Comparison -> Contrast -> however\\
3. Contingency -> Cause -> so\\
4. Contingency -> Pragmatic -> indeed\\
5. Expansion -> Alternative -> instead\\
6. Expansion -> Conjunction -> also\\
7. Expansion -> Instantiation -> for example\\
8. Expansion -> List -> and\\
9. Expansion -> Restatement -> specifically\\
10. Temporal -> Asynchronous -> before\\
11. Temporal -> Synchrony -> when\\
\textbf{ChatGPT:} Expansion -> List -> and\\
\midrule
\midrule
\textbf{Ground truth: Expansion.List} \\
\bottomrule
\end{tabular}
\vspace{-0.05in}
\caption{
Generation cases from three prompt templates for ChatGPT. The top one is the label-only template, the middle one is the label \& connectives template, and the bottom is the structural path template. 
}
\label{table:appendix_ChatGPT_Case_Study_1}
\vspace{-0.3cm}
\end{table}

\begin{table}[!htbp]
\small
\centering
\begin{tabular}{m{25em}}
\toprule
\textbf{Query Template:} Argument 1: Grinned Griffith Peck, a trader
in Shearson Lehman Hutton Inc.’s OTC department: "I tell you,
this market acts healthy". Argument 2: Around him, scores of
traders seemed to get a burst of energy. What is the relation
label between Argument 1 and Argument 2? Select from
the candidates.\\
1. Comparison.Concession\\
2. Comparison.Contrast\\
3. Contingency.Cause\\
4. Contingency.Pragmatic\\
5. Expansion.Alternative\\
6. Expansion.Conjunction\\
7. Expansion.Instantiation\\
8. Expansion.List\\
9. Expansion.Restatement\\
10. Temporal.Asynchronous\\
11. Temporal.Synchrony\\
\textbf{ChatGPT:} Contingency.Cause\\
\midrule
\midrule
\textbf{Query Template:} Argument 1: Grinned Griffith Peck, a trader
in Shearson Lehman Hutton Inc.’s OTC department: "I tell you,
this market acts healthy". Argument 2: Around him, scores of
traders seemed to get a burst of energy. What is the relation
and connective between Argument 1 and Argument 2? Select
from the candidates.\\
1. Comparison.Concession, if\\
2. Comparison.Contrast, however\\
3. Contingency.Cause, so\\
4. Contingency.Pragmatic, indeed\\
5. Expansion.Alternative, instead\\
6. Expansion.Conjunction, also\\
7. Expansion.Instantiation, for example\\
8. Expansion.List, and\\
9. Expansion.Restatement, specifically\\
10. Temporal.Asynchronous, before\\
11. Temporal.Synchrony, when\\
\textbf{ChatGPT:} Contingency.Cause, so\\
\midrule
\midrule
\textbf{Query Template:} Argument 1: Grinned Griffith Peck, a trader in Shearson Lehman Hutton Inc.’s OTC department: "I tell you, this market acts healthy". Argument 2: Around him, scores of
traders seemed to get a burst of energy. What is the relation
path between Argument 1 and Argument 2? Select from the
candidates.\\
1. Comparison -> Concession -> if\\
2. Comparison -> Contrast -> however\\
3. Contingency -> Cause -> so\\
4. Contingency -> Pragmatic -> indeed\\
5. Expansion -> Alternative -> instead\\
6. Expansion -> Conjunction -> also\\
7. Expansion -> Instantiation -> for example\\
8. Expansion -> List -> and\\
9. Expansion -> Restatement -> specifically\\
10. Temporal -> Asynchronous -> before\\
11. Temporal -> Synchrony -> when\\
\textbf{ChatGPT:} Temporal -> Synchrony -> when \\
\midrule
\midrule
\textbf{Ground truth: Temporal.Synchrony} \\
\bottomrule
\end{tabular}
\vspace{-0.05in}
\caption{
Generation cases from three prompt templates for ChatGPT. The top one is the label-only template, the middle one is the label \& connectives template, and the bottom is the structural path template. 
}
\label{table:appendix_ChatGPT_Case_Study_2}
\vspace{-0.3cm}
\end{table}

\end{document}